\documentclass{ecai} 
\usepackage{latexsym}
\usepackage{amssymb}
\usepackage{amsmath}
\usepackage{amsthm}
\usepackage{booktabs}
\usepackage{enumitem}
\usepackage{graphicx}
\usepackage{color}
\usepackage{multicol,multirow}
\usepackage{subfig} 

\usepackage{float}
\usepackage{pifont}% http://ctan.org/pkg/pifont
\newcommand{\cmark}{\ding{51}}%
\newcommand{\xmark}{\ding{55}}%
\newcommand{\BibTeX}{B\kern-.05em{\sc i\kern-.025em b}\kern-.08em\TeX}

\begin{document}

%%%%%%%%%%%%%%%%%%%%%%%%%%%%%%%%%%%%%%%%%%%%%%%%%%%%%%%%%%%%%%%%%%%%%%%%

\begin{frontmatter}

%%% Use this command to specify your submission number.
%%% In doubleblind mode, it will be printed on the first page.

% \paperid{464} 

%%% Use this command to specify the title of your paper.

\title{TimeMachine: A Time Series is Worth 4 Mambas for Long-term Forecasting}

%%% Use this combinations of commands to specify all authors of your 
%%% paper. Use \fnms{} and \snm{} to indicate everyone's first names 
%%% and surname. This will help the publisher with indexing the 
%%% proceedings. Please use a reasonable approximation in case your 
%%% name does not neatly split into "first names" and "surname".
%%% Specifying your ORCID digital identifier is optional. 
%%% Use the \thanks{} command to indicate one or more corresponding 
%%% authors and their email address(es). If so desired, you can specify
%%% author contributions using the \footnote{} command.

\author[A]{\fnms{Md Atik}~\snm{Ahamed}\thanks{Email: atikahamed@uky.edu}}
\author[A,B]{\fnms{Qiang}~\snm{Cheng}\thanks{Email: qiang.cheng@uky.edu, Corresponding Author.}}
\address{Department of Computer Science\textsuperscript{a}, Institute for Biomedical Informatics\textsuperscript{b}\\University of Kentucky}
% \address[B]{Institute for Biomedical Informatics, University of Kentucky}

%%% Use this environment to include an abstract of your paper.

\begin{abstract}
Long-term time-series forecasting remains challenging due to the difficulty in capturing long-term dependencies, achieving linear scalability, and maintaining computational efficiency. We introduce TimeMachine, an innovative model that leverages Mamba, a state-space model, to capture long-term dependencies in multivariate time series data while maintaining linear scalability and small memory footprints. TimeMachine exploits the unique properties of time series data to produce salient contextual cues at multi-scales and leverage an innovative integrated quadruple-Mamba architecture to unify the handling of channel-mixing and channel-independence situations, thus enabling effective selection of contents for prediction against global and local contexts at different scales. Experimentally, TimeMachine achieves superior performance in prediction accuracy, scalability, and memory efficiency, as extensively validated using benchmark datasets.\\ Code availability: \color{blue}{~\url{https://github.com/Atik-Ahamed/TimeMachine}}
\end{abstract}

\end{frontmatter}

%%%%%%%%%%%%%%%%%%%%%%%%%%%%%%%%%%%%%%%%%%%%%%%%%%%%%%%%%%%%%%%%%%%%%%%%

\section{Introduction}
Long-term time-series forecasting (LTSF) is essential in various tasks across diverse fields, such as weather forecasting, anomaly detection, and resource planning in energy, agriculture, industry, and defense. Although numerous approaches have been developed for LTSF, they typically can achieve only one or two desired properties such as capturing long-term dependencies in multivariate time series (MTS), linear scalability in the amount of model parameters with respect to data, and computational efficiency or applicability in edge computing. It is still challenging to achieve these desirable properties simultaneously. 

%% DISCUSSION OF LINEAR MODELS AND TRANSFORMER-BASED MODELS ==> SSM
Capturing long-term dependencies, which are generally abundant in MTS data, is pivotal to LTSF. While linear models such as DLinear \citep{dlinear} and TiDE \citep{tide}  achieve competitive performance with linear complexity and scalability, with accuracy on par with Transformer-based models, they usually rely on MLPs and linear projections that may not well capture long-range correlations \citep{bommasani2021opportunities}. Transformer-based models such as iTransformer \citep{liu2024itransformer}, PatchTST \citep{patchtst}, and Crossformer \citep{crossformer} have a strong ability to capture long-range dependencies and superior performance in LTSF accuracy, thanks to the self-attention mechanisms in Transformers \citep{vaswani2017attention}. However, they typically suffer from the quadratic complexity \citep{tide}, limiting their scalability and applicability, e.g., in edge computing.  

Recently, state-space models (SSMs) \citep{fu2022hungry,gu2023mamba,gu2021efficiently,gu2021combining,schiff2024caduceus} have emerged as powerful engines for sequence-based inference and have attracted growing research interest. These models are capable of inferring over very long sequences and exhibit distinctive properties, including the ability to capture long-range correlations with linear complexity and context-aware selectivity with hidden attention mechanisms \citep{gu2023mamba,hiddenattention}. SSMs have demonstrated great potential in various domains, including genomics \citep{gu2023mamba}, tabular learning \citep{mambatab}, graph data \citep{graphmamba}, and images \citep{umamba}, yet they remain unexplored for LTSF.

The under-utilization of SSMs in LTSF can be attributed to two main reasons. 
First, highly content- and context-selective SSMs have only been recently developed \citep{gu2023mamba}. Second, and more importantly, effectively representing the context in time series data remains a challenge. Many Transformer-based models, such as Autoformer \citep{autoformer} and Informer \citep{zhou2021informer}, regard each observation as a token in a sequence, while more recent models like PatchTST \citep{patchtst} and iTransformer \citep{liu2024itransformer} leverage patches of the time series as tokens. However, our empirical experiments on real-world MTS data suggest that 
directly utilizing SSMs for LTSF by using either observations or patches as tokens could hardly achieve performance comparable to Transformer-based models. Considering the particular characteristics of MTS data, it is essential to extract more salient contextual cues tailored to SSMs.

%% DISCUSSION OF CHANNEL INDEPENDENCE ADN MIXING
MTS data typically have many channels with each variate corresponding to a channel. Many models, such as Informer \citep{zhou2021informer}, FEDformer \citep{fedformer}, and Autoformer \citep{autoformer}, handle MTS data to extract useful representations in a channel-mixing way, where the MTS input is treated as a two-dimensional matrix whose size is the number of channels multiplied by the length of history. Nonetheless, recently a few works such as PatchTST \citep{patchtst} and TiDE \citep{tide} have shown that a channel-independence way for handling MTS may achieve state-of-the-art (SOTA) accuracy, where each channel is input to the model as a one-dimensional vector independent of the other channels. We believe that these two ways of handling LTSF need to be adopted as per the characteristics of the MTS data, rather than using a one-size-fits-all approach. When there are strong between-channel correlations, channel mixing usually can help capture such dependencies; otherwise, channel independence is a more sensible choice. Therefore, it is necessary to design a unified architecture applicable to both channel-mixing and channel-independence scenarios. 

%% MULTI-SCALE , GLOBAL AND LOCAL CONTEXTS
Moreover, time series data exhibit a unique property -- Temporal relations are largely preserved after downsampling into two sub-sequences. Few methods such as Scinet \citep{scinet} have explored this property in designing their models; however, it is under-utilized in other approaches. Due to the high redundancy of MTS values at consecutive time points, directly using time points as tokens may have redundant values obscure context-based selection and, more importantly, overlook long-range dependencies. Rather than relying on individual time points, using patches may provide contextual clues within each time window of a patch length. However, a pre-defined small patch length only provides contexts at a fixed temporal or frequency resolution, whereas long-range contexts may span different patches. To best capture long-range dependencies, it is sensible to supply multi-scale contexts and, at each scale, automatically produce global-level tokens as contexts similar to iTransformer \citep{liu2024itransformer} that tokenizes the whole look-back window. Further, while models like Transformer and the selective SSMs \citep{gu2023mamba} have the ability to select sub-token contents, such ability is limited in the channel-independence case, for which local contexts need to be enhanced when leveraging SSMs for LTSF.  

In this paper, we introduce a novel approach that effectively captures long-range dependencies in time series data by providing rich multi-scale contexts and particularly enhancing local contexts in the channel-independence situation. Our model, built upon a selective scan SSM called Mamba \citep{gu2023mamba}, serves as a core inference engine with a strong ability to capture long-range dependencies in MTS data while maintaining linear scalability and small memory footprints. The proposed model exploits the unique property of time series data in a bottom-up manner by producing contextual cues at two scales through consecutive resolution reduction or downsampling using linear mapping. The first level operates at a high resolution, while the second level works at a low resolution. At each level, we employ two Mamba modules to glean contextual cues from global perspectives for the channel-mixing case and from both global and local perspectives for the channel-independence case.
 
In summary, our major contributions are threefold:
\begin{itemize}
    \item We develop an innovative model called TimeMachine that is the first to leverage purely SSM modules to capture long-term dependencies in multivariate time series data for context-aware prediction, with linear scalability and small memory footprints superior or comparable to linear models.
    \item Our model constitutes an innovative architecture that unifies the handling of channel-mixing and channel-independence situations with four SSM modules, exploiting potential between-channel correlations. Moreover, our model can effectively select contents for prediction against global and local contextual information, at different scales in the MTS data. 
    \item Experimentally, TimeMachine achieves superior performance in prediction accuracy, scalability, and memory efficiency. We extensively validate the model using standard benchmark datasets and perform rigorous ablation studies to demonstrate its effectiveness. 
\end{itemize}

\section{Related Works}
Numerous methods for LTSF have been proposed, which can be grouped into three main categories: non-Transformer-based supervised approaches, Transformer-based supervised learning models, and self-supervised representation learning models. \\

\noindent{\bf{Non-Transformer-based Supervised Approaches}}
include classical methods like ARIMA, VARMAX, GARCH \citep{box2015time}, and RNN \citep{hochreiter1997long}, as well as deep learning-based methods that achieve SOTA performance using multi-layer perceptrons (MLPs) and convolutional neural networks (CNNs). 
MLP-based models, such as DLinear \citep{dlinear}, TiDE \citep{tide}, and RLinear \citep{rlinear}, leverage the simplicity of linear structures to 
achieve complexity and scalability. 
CNN-based methods, such as TimesNet \citep{timesnet} and Scinet \citep{scinet}, utilize convolutional filters to extract valuable temporal features and model complex temporal dynamics. These approaches exhibit highly competitive performance, often comparable to or even occasionally outperforming more sophisticated Transformer-based models. 

\noindent{\bf{Transformer-based Supervised Learning Methods,}}
%Transformers  \citep{vaswani2017attention} have been predominantly used for building foundation models \citep{bommasani2021opportunities}. Recently, they have been popular for LTSF, leading to superior accuracy. Representative methods include: iTransformer \citep{liu2024itransformer}, PatchTST \citep{patchtst}, Crossformer \citep{crossformer}, FEDformer \citep{fedformer}, stationary \citep{nonstationary} and  autoformer \citep{autoformer}.  These methods have different ways to convert WTS to token sequences and leverage the self-attention mechanism of Transformers to discover dependencies between arbitrary time steps in a time series, making them particularly effective for modeling complex temporal relationships.  Another important property of these methods is to exploit  Transformers' ability to process data in parallel, thus enabling their discovery of long-term dependency and linear scalability.  Despite distinctive advantages, these methods typically have quadratic complexity due to the point-wise correlations in the self-attention mechanisms. While efforts have been made to reduce the complexity of Transformers, the resulting methods typically have degraded accuracy. 
such as iTransformer \citep{liu2024itransformer}, PatchTST \citep{patchtst}, Crossformer \citep{crossformer}, FEDformer \citep{fedformer}, stationary \citep{nonstationary}, Flowformer \citep{wu2022flowformer}, and Autoformer \citep{autoformer}, have gained popularity for LTSF due to their superior accuracy. These methods convert time series to token sequences and leverage the self-attention mechanism to discover dependencies between arbitrary time steps, making them particularly effective for modeling complex temporal relationships. They may also exploit Transformers' ability to process data in parallel, enabling long-term dependency discovery sometimes with even linear scalability. Despite their distinctive advantages, these methods typically have quadratic time and memory complexity due to point-wise correlations in self-attention mechanisms. 

\noindent{\bf{Self-Supervised Representation Learning Models:}}
Self-supervised learning has been leveraged to learn useful representations of MTS for downstream tasks, using non-Transformer-based models for time series \citep{yang2022unsupervised, franceschi2019unsupervised, tonekaboni2021unsupervised, yue2022ts2vec}, and Transformer-based models such as time series Transformer (TST)  and TS-TCC \citep{zerveas2021transformer, eldele2021time, trirat2024universal}.
Currently, Transformer-based self-supervised models have not yet achieved performance on par with supervised learning approaches \citep{trirat2024universal}. This paper focuses on LTSF in a supervised learning setting. 
\section{Proposed Method}
\begin{figure*}[t]
    \centering
    \includegraphics[width=0.9\linewidth]{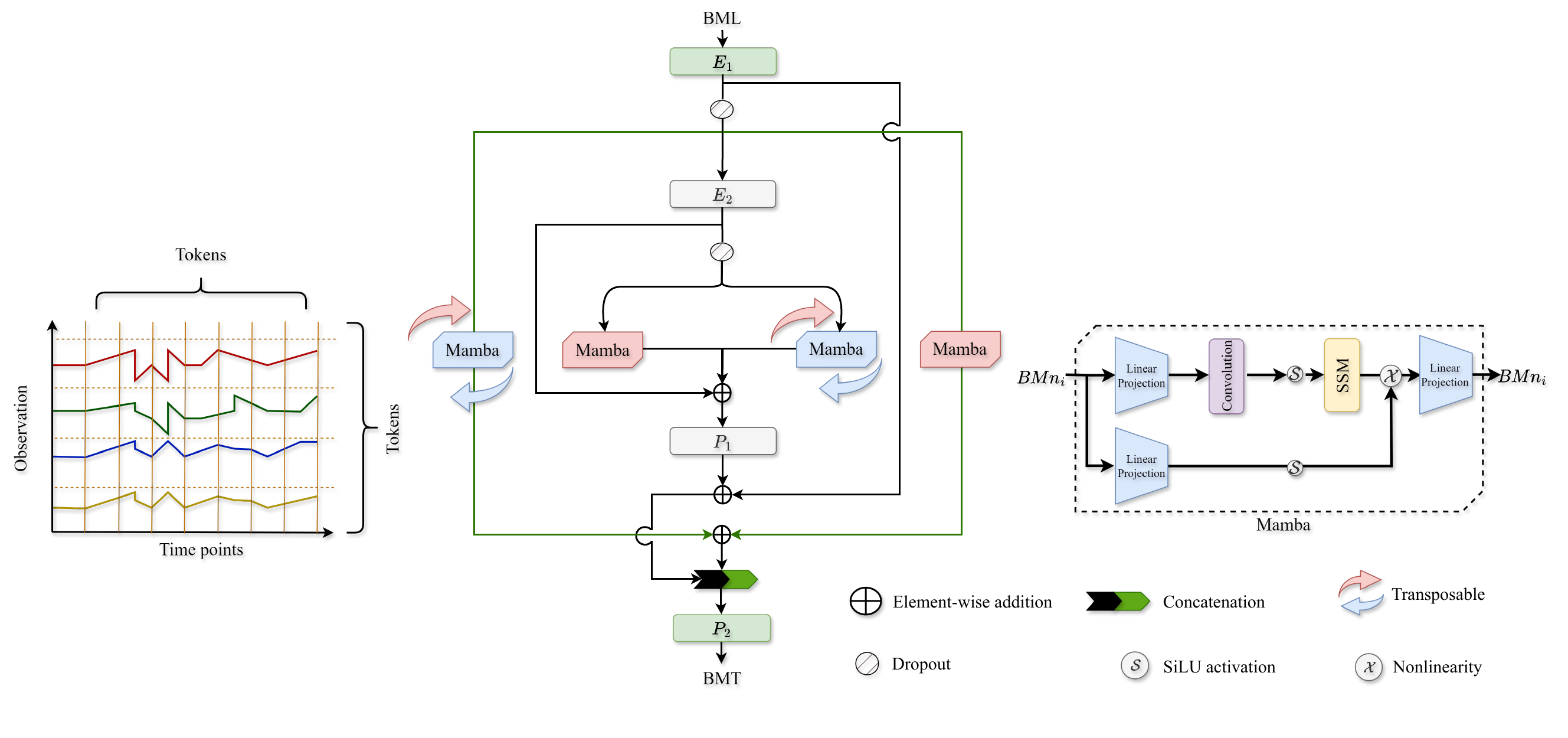}
    \caption{Schematic diagram of our proposed model, TimeMachine. Our method incorporates a configuration of four Mambas, with two specialized Mambas capable of processing the transposed tensor data in each branch. On the left, an example of the MTS is depicted, while the right side shows a detailed view of a Mamba's structure. Mambas are capable of accepting an input of shape $BMn_i$ while providing the output of the same shape, where $i\in\{1,2\}$ in our method.}
    \vspace{1mm}
    \label{fig:method}
\end{figure*}
In this section, we describe each component of our proposed architecture and how we use our model to solve the LTSF problem. Assume a collection of MTS samples is given, denoted by dataset $\mathcal{D}$, which comprises an input sequence $\mathbf{x} = [x_1,\dots,x_L]$, with each $x_t \in {\mathcal{R}}^M$ representing a vector of $M$ channels at time point $t$. For matrices, we use bold font; for scalars and vectors, we use regular (non-bold) letters. The sequence length $L$ is also known as the look-back window. The goal is to predict $T$ future values, denoted by $[x_{L+1},\dots,x_{L+T}]$. The architecture of our proposed model, referred to as TimeMachine, is depicted in Figure~\ref{fig:method}. The pillars of this architecture consist of four Mambas, which are employed in an integrated way to tap contextual cues from MTS. This design choice enables us to harness Mamba's robust capabilities of inferring sequential data for LTSF.\\ 

\noindent{\bf{Normalization:}} Before feeding the  data to our model, we normalize the original MTS $\mathbf{x}$ into $\mathbf{x}^{0} = [\mathbf{x}^{(0)}_1, \cdots, \mathbf{x}^{(0)}_L] \in {\mathcal{R}^{M \times L}}$, via  
%\begin{equation}
    $\mathbf{x}^{(0)} = Normalize(\mathbf{x})$.  
%\end{equation}
Here, $Normalize(\cdot)$ represents a normalization operation with two different options. The first is to use the reversible instance normalization (RevIN)~\citep{kim2022reversible}, which is also adopted in PatchTST~\citep{patchtst}. The second option is to employ regular $Z$-score normalization:
%\begin{equation}
%\label{eq:z_score}
$x^{(0)}_{i, j} = ({x_{i,j} - mean(x_{i,j})) / \sigma_j}$, 
%\end{equation}
where $\sigma_j$ is the standard\ deviation\ for\ channel $j$, with $j=1, \cdots, M$. Empirically we find that RevIN is often more helpful compared to Z-score. Apart from normalizing the data in the forward pass of our approach, in experiments, we also follow the standardization process of the data when compared with baseline methods.\\

\noindent{\bf{Channel Mixing vs. Channel Independence:}}
Our model can handle both channel independence and channel mixing cases. In channel independence, each channel is processed independently by our model, while in channel mixing, the MTS sequence is processed with multiple channels combined throughout our architecture. Regardless of the case, our model accepts input of the shape $BML$, where $B$ is batch size, and produces the desired output of the shape $BMT$, eliminating the need for additional manual pre-processing. 

Channel independence has been proven effective in reducing overfitting by PatchTST~\citep{patchtst}. We found this strategy helpful for datasets with a smaller number of channels. However, we observe for datasets with a number of channels comparable to the look-back, channel mixing is more effective in capturing the correlations among channels and reaching the desired minimum loss during training. 

Our architecture is robust and versatile, capable of benefiting from potentially strong inter-channel correlations in channel-mixing case and exploiting independence in channel-independence case. When dealing with channel independence, we reshape the input from $BML$ to $(B\times M)1 L$ after the normalization step. The reshaped input is then processed throughout the network and later merged to provide an output shape of $BMT$. In contrast, for channel mixing, no reshaping is necessary. The channels are kept together and processed throughout the network.  \\

\noindent{\bf{Embedded Representations:}}
Before processing the input sequence with Mambas, we provide two-stage embedded representations of the input sequence with length $L$ by $E_1$ and $E_2$:
\begin{equation}
    \mathbf{x}^{(1)} = E_1 (\mathbf{x}^{(0)}), \quad  \mathbf{x}^{(2)} = E_2 (DO(\mathbf{x}^{(1)})),
\end{equation}
where $DO$ stands for the dropout operation, and the embedding operations $E_1\!:$  $\mathbb{R}^{M \times L} \rightarrow \mathbb{R}^{M \times n_1}$ and $E_2\!:$  $\mathbb{R}^{M \times n_1} \rightarrow \mathbb{R}^{M \times n_2}$ are achieved through MLPs. Thus, for the channel mixing case, the batch-formed tensors will have the following changes in size:
%\begin{equation}
%\label{eq:emb}
%\begin{split}
   $ BMn_1 \leftarrow E_1(BML)$, and $BMn_2 \leftarrow E_2(BMn_1)$.
%\end{split}
%\end{equation}
This enables us to deal with the fixed-length tokens of $n_1$ and $n_2$ regardless of the variable input sequence length $L$, and both $n_1$ and $n_2$ are configured to take values from the set $\{512,256,128,64,32\}$ satisfying $n_1>n_2$. Since MLPs are fully connected, we introduce dropouts to reduce overfitting. Although we have the linear mappings (MLPs) before Mambas, the performance of our model does not heavily rely on them, as demonstrated with the ablation study (see Section~\ref{sec:ablation}). \\

\noindent{\bf{Integrated Quadruple Mambas:}} With the two processed embedded representations from $E_1,E_2$, we can now learn more comprehensive representations by leveraging Mamba, a type of SSM with selective scan ability. At each embedding level, we employ a pair of Mambas to capture long-term dependencies within the look-back samples and provide sufficient local contexts. Denote the input to one of the four Mamba blocks by $u$, which is either $DO(\mathbf{x}^{(1)})$ obtained after $E_1$ and the subsequent dropout layer for the two outer Mambas, or $DO(\mathbf{x}^{(2)})$ obtained after $E_2$ and the subsequent dropout layer for the two inner Mambas (Figure~\ref{fig:method}). The input tensors may be reshaped per channel mixing or channel independence cases as described. 

Inside a Mamba block, two fully-connected layers in two branches calculate linear projections. The output of the linear mapping in the first branch passes through a 1D causal convolution and SiLU activation ${\mathcal{S}}(\cdot)$~\citep{silu}, then a structured SSM. The continuous-time SSM maps an input function or sequence $u(t)$ to output $v(t)$ through a latent state $h(t)$: 
\begin{equation}
    \label{eq:CT-SSM}
    dh(t)/dt = A \ h(t) + B \  u(t), \quad v(t) = C \ h(t), 
\end{equation}
where $h(t)$ is $N$-dimensional, with $N$ also known as a {\sl{state expansion factor}}, $u(t)$ is $D$-dimensional, with $D$ being the {\sl{dimension factor}} for an input token, $v(t)$ is an output of dimension $D$, and $A$, $B$, and $C$ are coefficient matrices of proper sizes. 
This dynamic system induces a discrete SSM governing state evolution and outputs given the input token sequence through time sampling at $\{ k \Delta \}$. Here, $\Delta$ is a time interval for discretizing the dynamic system. In particular, Mamba makes $\Delta$ a function of the input, and hence so do the model coefficients (A, B, C) and hidden state, thereby adapting the model dynamics to input and enhancing context selectivity. Consequently, this discrete SSM is 
\begin{equation}
    \label{eq:DT-SSM}
    h_k = \bar{A} \ h_{k-1} + \bar{B} \ u_{k}, \quad v_k = C \ h_{k}, 
\end{equation}
where $h_k$, $u_k$, and $v_k$ are respectively samples of $h(t)$, $u(t)$, and $v(t)$ at time $k \Delta$, 
\begin{equation}
    \bar{A} = \exp(\Delta A), \quad 
    \bar{B} = (\Delta A)^{-1} (\exp(\Delta A) - I) \Delta B.    
\end{equation}
For SSMs, diagonal $A$ is often used. Mamba makes $B$, $C$, and $\Delta$ linear time-varying functions dependent on the input. In particular, for a token $u$,  
%\begin{equation}
%\begin{split}
$B,C  \leftarrow Linear_N(u)$,  and $\Delta  \leftarrow softplus(parameter + Linear_D(Linear_1 (u)))$, 
%\end{split}
%\end{equation}
where $Linear_p (u)$ is a linear projection to a $p$-dimensional space, and $softplus$ activation function. 
Furthermore, Mamba also has an option to expand the model dimension factor $D$ by a controllable dimension expansion factor $E$.  
Such coefficient matrices enable context and input selectivity properties \citep{gu2023mamba} to selectively propagate or forget information along the input token sequence based on the current token. Subsequently, the SSM output is multiplicatively modulated with the output from the second branch before another fully connected projection.  The second branch simply consists of a linear mapping followed by a SiLU. 

Processed embedded representation with tensor size $BMn_1$ is transformed by outer Mambas, while that with $BMn_2$ is transformed by inner Mambas, as depicted in Figure~\ref{fig:method}. For the channel-mixing case, the whole univariate sequence of each channel is used as a token with dimension factor $n_2$ for the inner Mambas. The outputs from the left-side and right-side inner Mambas, $v_{L, k}, v_{R,k} \in {\mathcal{R}^{n_2}}$, are element-wise added with $x^{(2)}_k$ to obtain $x^{(3)}_k$  
for the $k$-th token, $k=1, \cdots, M$. That is, by denoting $\mathbf{v}_{L} = [v_{L, 1}, \cdots, v_{L, M}]^T \in {\mathcal{R}^{M \times n_2}}$ and similarly $\mathbf{v}_{R} \in {\mathcal{R}^{M \times n_2}}$, we have  
$\mathbf{x}^{(3)} = \mathbf{v}_{L} \bigoplus \mathbf{v}_{R} \bigoplus \mathbf{x}^{(2)}$, with $\bigoplus$ being element-wise addition. Then, $\mathbf{x}^{(3)}$ is linearly mapped to $\mathbf{x}^{(4)}$ with $P_1: \mathbf{x}^{(3)} \rightarrow \mathbf{x}^{(4)} \in {\mathcal{R}^{M \times n_1}}$.  
Similarly, the outputs from the outer Mambas, $v^{*}_{L, k}, v^{*}_{R,k} \in {\mathcal{R}^{n_1}}$ are element-wise added to obtain
$\mathbf{x}^{(5)} \in {\mathcal{R}^{M \times n_1}}$. %, that is, $x^{(5)} = v^{*}_L \bigoplus v^{*}_R$.

For the channel independence case, the input is reshaped, $BML\mapsto(B\times M)1L$, and the embedded representations become 
$(B\times M)1 n_1$ and $(B\times M)1 n_2$. Here, the batch size becomes $B\times M$, and we regard each sequence of length $L$ independent from each other. One Mamba in each pair of outer Mambas or inner Mambas considers the input dimension as 1 and the token length as $n_1$ or $n_2$, while the other Mamba learns with input dimension $n_2$ or $n_1$ and token length 1. This design enables learning both global context and local context simultaneously. 
The outer and inner pairs of Mambas will extract salient features and context cues at fine and coarse scales with high- and low-resolution, respectively. 

Channel mixing is performed when the datasets contain a significantly large number of channels, in particular, when the look-back $L$ is comparable to the channel number $M$, taking the whole sequence as a token to better provide context cues. All four Mambas are used to capture the global context of the sequences at different scales and learn from the channel correlations. This helps stabilize the training and reduce overfitting with large $M$. To switch between the channel-independence and channel-mixing cases, the input sequence is simply transposed, with one Mamba in the outer Mamba pair and one in the inner pair processing the transposed input, as demonstrated in Figure~\ref{fig:method}.
These integrated Mamba blocks empower our model for content-dependent feature extraction and reasoning with long-range dependencies and feature interactions.  \\

\noindent{\bf{Output Projection:}}
After receiving the output tokens from the Mambas, our goal is to project these tokens to generate predictions with the desired sequence length. To accomplish this task, we utilize two MLPs, $P_1$ and $P_2$, which output $n_1$ and $T$ time points, respectively, with each point having $M$ channels. Specifically, projector $P_1$ performs a mapping $\mathcal{R}^{M \times n_2} \rightarrow \mathcal{R}^{M \times n_1}$, as discussed above for obtaining $\mathbf{x}^{(4)}$. Subsequently, projector $P_2$ performs a mapping $\mathbb{R}^{M \times 2n_1} \rightarrow \mathbb{R}^{M \times T}$, transforming the concatenated output from the Mambas into the final predictions. The use of a two-stage output projection via $P_1$ and $P_2$ symmetrically aligns with the two-stage embedded representation obtained through $E_1$ and $E_2$.

In addition to the token transformation, we also employ residual connections. One residual connection is added before $P_1$, and another is added after $P_1$. The effectiveness of these residual connections is verified by experimental results (see Supplementary Table~\ref{tab:res_connections}). Residual connections are demonstrated by arrows and element-wise addition in our method (Figure~\ref{fig:method}). 

To retain the information of both outer and inner pairs of Mambas we concatenate their representations before processing via $P_2$. In summary, we concatenate the outputs of the four Mambas with a skip connection to have $\mathbf{x}^{(6)}  = \mathbf{x}^{(5)} \| (\mathbf{x}^{(4)} \bigoplus \mathbf{x}^{(1)})$,  where $\|$ denotes concatenation. Finally, the output $y$ is obtained by applying $P_2$ to $\mathbf{x}^{(6)}$, i.e., $y = P_2 (\mathbf{x}^{(6)})$.

\section{Result Analysis}
\label{sec:result}
In this segment, we present the main results of our experiments on widely recognized benchmark datasets for long-term MTS forecasting. We also conduct extensive ablation studies to demonstrate the effectiveness of each component of our method.
\subsection{Datasets}
We evaluate our model on seven benchmark datasets extensively used for LTSF: Weather, Traffic, Electricity, and four ETT datasets (ETTh1, ETTh2, ETTm1, ETTm2). Table~\ref{tab:datasets} illustrates the relevant statistics of these datasets, highlighting that the Traffic and Electricity datasets notably large, with 862 and 321 channels, respectively, and tens of thousands of temporal points in each sequence. More details on these datasets can be found in~\citet{autoformer,zhou2021informer}. Focusing on long-term forecasting, we exclude the ILI dataset, which has a shorter temporal horizon, similar to~\citet{tide}.
We demonstrate the superiority of our model in two parts: quantitative (main results) and qualitative results. 
For a fair comparison, we used the code from PatchTST~\citep{patchtst} \footnote{https://github.com/yuqinie98/PatchTST} and iTransformer~\citep{liu2024itransformer} \footnote{https://github.com/thuml/iTransformer} including normalization and evaluation protocol used by them, and we took the results for the baseline methods from  iTransformer~\citep{liu2024itransformer}.
\begin{table}[t]
\centering
\caption{Overview of the characteristics of used benchmarking datasets. Time points illustrate the total length of each dataset.}
\resizebox{\linewidth}{!}{
\begin{tabular}{l|c|c|c}
\toprule
Dataset ($\mathcal{D}$)&Channels (M)&Time Points&Frequency \\
\midrule
Weather & 21 & 52696 & 10 Minutes \\
Traffic & 862 & 17544 & Hourly \\
Electricity & 321 & 26304 & Hourly \\
ETTh1   & 7 & 17420 & Hourly \\
ETTh2   & 7 & 17420 & Hourly \\
ETTm1   & 7 & 69680 & 15 Minutes \\
ETTm2   & 7 & 69680 & 15 Minutes \\
\bottomrule
\end{tabular}
}
\label{tab:datasets}
\end{table}

\subsection{Experimental Environment}
\label{subsection:environment}
All experiments were conducted using the PyTorch framework~\citep{paszke2019pytorch} with NVIDIA 4XV100 GPUs (32GB each). The model was optimized using the ADAM algorithm~\citep{kingma2014adam} with $L_2$ loss. The batch size varied depending on the dataset, but the training was consistently set to 100 epochs. We measure the prediction errors using mean square error (MSE) and mean absolute error (MAE) metrics, where smaller values indicate better prediction accuracy. \\

{\bf{Baseline Models:}} We compared our model, TimeMachine, with 11 SOTA models, including iTransformer~\citep{liu2024itransformer}, PatchTST~\citep{patchtst}, DLinear~\citep{dlinear}, RLinear~\citep{rlinear}, Autoformer~\citep{autoformer}, Crossformer~\citep{crossformer}, TiDE~\citep{tide}, Scinet~\citep{scinet}, TimesNet~\citep{timesnet}, FEDformer~\citep{fedformer}, and Stationary~\citep{nonstationary}.  Although another variant of SSMs, namely S4~\citep{gu2021efficiently}, exists, we did not include it in our comparison because TiDE~\citep{tide} has already demonstrated superior performance over S4. Similarly, as Flowformer~\citep{wu2022flowformer} is not as competitive as iTransformer and other Transformer-based models, following~\citep{liu2024itransformer}, we did not include it. 

\subsection{Quantitative Results}
We demonstrate TimeMachine's performance in supervised long-term forecasting tasks in Table~\ref{tab:result}. Following the protocol used in iTransformer~\citep{liu2024itransformer}, we set all baselines fixed with $L=96$ and $T=\{96, 192, 336, 720\}$, including our method. For all results achieved by our model, we utilized the training-related values mentioned in Subsection~\ref{subsection:environment}. In addition to the training hyperparameters, we set default values for all Mambas: Dimension factor $D=256$, local convolutional width $=2$, and state expand factor $N= 1$. We provide an experimental justification for these parameters in Section~\ref{sec:ablation}. Table~\ref{tab:result} clearly shows that our method demonstrates superior performance compared to all the strong baselines in almost all datasets. Moreover, iTransformer~\citep{liu2024itransformer} has significantly better performance than other baselines on the Traffic and Electricity datasets, which contain a large number of channels. Our method also demonstrates comparable or superior performance on these two datasets, outperforming the existing strong baselines by a large margin. This demonstrates the effectiveness of our method in handling LTSF tasks with varying number of channels and datasets. 

In addition to Table~\ref{tab:result}, we conducted experiments with TimeMachine using different look-back windows $L=\{192,336,720\}$. Table~\ref{tab:result_long_l} and Supplementary Table~\ref{tab:result_long_l_s}, demonstrate TimeMachine's performance under these settings. An examination of these tables reveals that the implementation of extended look-back windows markedly enhances the performance of our method across the majority of the datasets examined. This also demonstrates TimeMachine's capability for handling longer look-back windows while maintaining consistent performance.

\begin{table*}[t]
\centering
\vspace{4mm}
\caption{Results in MSE and MAE (the lower the better) for the long-term forecasting task. We compare extensively with baselines under different prediction lengths, $T=\{96,192,336,720\}$ following the setting of iTransformer~\citep{liu2024itransformer}. The length of the input sequence ($L$) is set to 96 for all baselines. The best results are in \textbf{bold} and the second best are \underline{underlined}.}
\label{tab:result}
\setlength{\tabcolsep}{2pt}
\medskip
\resizebox{\linewidth}{!}{
\begin{tabular}{lc|cc|cc|cc|cc|cc|cc|cc|cc|cc|cc|cc|cc} 
\toprule
          
\multicolumn{2}{c}{Methods$\rightarrow$}&\multicolumn{2}{c|}{TimeMachine} & \multicolumn{2}{c|}{iTransformer} & \multicolumn{2}{c|}{RLinear} & \multicolumn{2}{c|}{PatchTST} & \multicolumn{2}{c|}{Crossformer} & \multicolumn{2}{c|}{TiDE} & \multicolumn{2}{c|}{TimesNet}  & \multicolumn{2}{c|}{DLinear} & \multicolumn{2}{c|}{SCINet} & \multicolumn{2}{c|}{FEDformer}& \multicolumn{2}{c|}{Stationary}& \multicolumn{2}{c}{Autoformer}\\ 
\midrule
$\mathcal{D}$& $T$ & MSE & MAE & MSE & MAE & MSE & MAE & MSE & MAE & MSE & MAE & MSE & MAE & MSE & MAE & MSE & MAE & MSE & MAE & MSE & MAE & MSE & MAE & MSE & MAE\\
\midrule
\multirow{4}{*}{\rotatebox{90}{Weather}}&96&\underline{0.164}  &\bf0.208&0.174&\underline{0.214}&0.192&0.232&0.177&0.218&\bf0.158&0.230&0.202&0.261&0.172&0.220&0.196&0.255&0.221&0.306&0.217&0.296&0.173&0.223&0.266&0.336\\
&192&\underline{0.211}&  \bf0.250&0.221&\underline{0.254}&0.240&0.271&0.225&0.259&\bf0.206&0.277&0.242&0.298&0.219&0.261&0.237&0.296&0.261&0.340&0.276&0.336&0.245&0.285&0.307&0.367\\
&336&\bf0.256&\bf0.290&0.278&\underline{0.296}&0.292&0.307&0.278&0.297&\underline{0.272}&0.335&0.287&0.335&0.280&0.306&0.283&0.335&0.309&0.378&0.339&0.380&0.321&0.338&0.359&0.395\\
&720&\bf0.342&  \bf0.343&0.358&\underline{0.349}&0.364&0.353&0.354&0.348&0.398&0.418&\underline{0.351}&0.386&0.365&0.359&0.345&0.381&0.377&0.427&0.403&0.428&0.414&0.410&0.419&0.428\\
\midrule
\multirow{4}{*}{\rotatebox{90}{Traffic}} &96&\underline{0.397}&\bf0.268&\bf0.395&\bf0.268&0.649&0.389&0.544&0.359&0.522&\underline{0.290}&0.805&0.493&0.593&0.321&0.650&0.396&0.788&0.499&0.587&0.366&0.612&0.338&0.613&0.388\\
& 192&\bf0.417&\bf0.274&\bf0.417&\underline{0.276}&0.601&0.366&0.540&0.354&\underline{0.530}&0.293&0.756&0.474&0.617&0.336&0.598&0.370&0.789&0.505&0.604&0.373&0.613&0.340&0.616&0.382\\
& 336&\bf0.433&\bf0.281&\bf0.433&\underline{0.283}&0.609&0.369&\underline{0.551}&0.358&0.558&0.305&0.762&0.477&0.629&0.336&0.605&0.373&0.797&0.508&0.621&0.383&0.618&0.328&0.622&0.337\\
& 720&\bf0.467& \bf0.300&\bf0.467&\underline{0.302}&0.647&0.387&\underline{0.586}&0.375&0.589&0.328&0.719&0.449&0.640&0.350&0.645&0.394&0.841&0.523&0.626&0.382&0.653&0.355&0.660&0.408\\
\midrule
\multirow{4}{*}{\rotatebox{90}{Electricity}}&96&\bf0.142&\bf0.236&\underline{0.148}&\underline{0.240}&0.201&0.281&0.195&0.285&0.219&0.314&0.237&0.329&0.168&0.272&0.197&0.282&0.247&0.345&0.193&0.308&0.169&0.273&0.201&0.317\\
&192&\bf0.158&\bf0.250&\underline{0.162}&\underline{0.253}&0.201&0.283&0.199&0.289&0.231&0.322&0.236&0.330&0.184&0.289&0.196&0.285&0.257&0.355&0.201&0.315&0.182&0.286&0.222&0.334\\
&336&\bf0.172&\bf0.268&\underline{0.178}&\underline{0.269}&0.215&0.298&0.215&0.305&0.246&0.337&0.249&0.344&0.198&0.300&0.209&0.301&0.269&0.369&0.214&0.329&0.200&0.304&0.231&0.338\\
&720&\bf0.207&\bf0.298&0.225&\underline{0.317}&0.257&0.331&0.256&0.337&0.280&0.363&0.284&0.373&\underline{0.220}&0.320&0.245&0.333&0.299&0.390&0.246&0.355&0.222&0.321&0.254&0.361\\
\midrule
\multirow{4}{*}{\rotatebox{90}{ETTh1}}&96&\bf0.364&\bf0.387&0.386&0.405&0.386&\underline{0.395}&0.414&0.419&0.423&0.448&0.479&0.464&0.384&0.402&0.386&0.400&0.654&0.599&\underline{0.376}&0.419&0.513&0.491&0.449&0.459\\
& 192 &\bf0.415&\bf0.416&0.441&0.436&0.437&\underline{0.424}&0.460&0.445&0.471&0.474&0.525&0.492&0.436&0.429&0.437&0.432&0.719&0.631&\underline{0.420}&0.448&0.534&0.504&0.500&0.482\\
& 336 &\bf0.429&\bf0.421&0.487&0.458&0.479&\underline{0.446}&0.501&0.466&0.570&0.546&0.565&0.515&0.491&0.469&0.481&0.459&0.778&0.659&\underline{0.459}&0.465&0.588&0.535&0.521&0.496\\
& 720 &\bf0.458&\bf0.453&0.503&0.491&\underline{0.481}&\underline{0.470}&0.500&0.488&0.653&0.621&0.594&0.558&0.521&0.500&0.519&0.516&0.836&0.699&0.506&0.507&0.643&0.616&0.514&0.512\\
\midrule
\multirow{4}{*}{\rotatebox{90}{ETTh2}}&96&\bf0.275&\bf0.334&0.297&0.349&\underline{0.288}&\underline{0.338}&0.302&0.348&0.745&0.584&0.400&0.440&0.340&0.374&0.333&0.387&0.707&0.621&0.358&0.397&0.476&0.458&0.346&0.388\\
&192&\bf0.349&\bf0.381&0.380&0.400&\underline{0.374}&\underline{0.390}&0.388&0.400&0.877&0.656&0.528&0.509&0.402&0.414&0.477&0.476&0.860&0.689&0.429&0.439&0.512&0.493&0.456&0.452\\
&336&\bf0.340&\bf0.381&0.428&0.432&\underline{0.415}&\underline{0.426}&0.426&0.433&1.043&0.731&0.643&0.571&0.452&0.452&0.594&0.541&1.000&0.744&0.496&0.487&0.552&0.551&0.482&0.486\\
& 720&\bf0.411&\bf0.433&0.427&0.445&\underline{0.420}&\underline{0.440}&0.431&0.446&1.104&0.763&0.874&0.679&0.462&0.468&0.831&0.657&1.249&0.838&0.463&0.474&0.562&0.560&0.515&0.511\\
\midrule
\multirow{4}{*}{\rotatebox{90}{ETTm1}} & 96&\bf0.317&\bf0.355&0.334&0.368&0.355&0.376&\underline{0.329}&\underline{0.367}&0.404&0.426&0.364&0.387&0.338&0.375&0.345&0.372&0.418&0.438&0.379&0.419&0.386&0.398&0.505&0.475\\
&192&\bf0.357&\bf0.378&0.377&0.391&0.391&0.392&\underline{0.367}&\underline{0.385}&0.450&0.451&0.398&0.404&0.374&0.387&0.380&0.389&0.439&0.450&0.426&0.441&0.459&0.444&0.553&0.496\\
&336&\bf0.379&\bf0.399&0.426&0.420&0.424&0.415&\underline{0.399}&\underline{0.410}&0.532&0.515&0.428&0.425&0.410&0.411&0.413&0.413&0.490&0.485&0.445&0.459&0.495&0.464&0.621&0.537\\
&720&\bf0.445&\bf0.436&0.491&0.459&0.487&0.450&\underline{0.454}&\underline{0.439}&0.666&0.589&0.487&0.461&0.478&0.450&0.474&0.453&0.595&0.550&0.543&0.490&0.585&0.516&0.671&0.561\\
\midrule
\multirow{4}{*}{\rotatebox{90}{ETTm2}} & 96&\bf0.175&\bf0.256&\underline{0.180}&0.264&0.182&0.265&\bf0.175&\underline{0.259}&0.287&0.366&0.207&0.305&0.187&0.267&0.193&0.292&0.286&0.377&0.203&0.287&0.192&0.274&0.255&0.339\\
& 192&\bf0.239&\bf0.299&0.250&0.309&0.246&0.304&\underline{0.241}&\underline{0.302}&0.414&0.492&0.290&0.364&0.249&0.309&0.284&0.362&0.399&0.445&0.269&0.328&0.280&0.339&0.281&0.340\\
& 336&\bf0.287&\bf0.332&0.311&0.348&0.307&\underline{0.342}&\underline{0.305}&0.343&0.597&0.542&0.377&0.422&0.321&0.351&0.369&0.427&0.637&0.591&0.325&0.366&0.334&0.361&0.339&0.372\\
& 720&\bf0.371&\bf0.385&0.412&0.407&0.407&\underline{0.398}&\underline{0.402}&0.400&1.730&1.042&0.558&0.524&0.408&0.403&0.554&0.522&0.960&0.735&0.421&0.415&0.417&0.413&0.433&0.432\\
\bottomrule
\end{tabular}
% \vspace{2mm}
}
\end{table*}
\begin{table}[t]
\centering
\caption{Results for the long-term forecasting task with varying $L=\{192,336,720\}$ and $T=\{96,192,336,720\}$}
\label{tab:result_long_l}
\setlength{\tabcolsep}{2pt}
\medskip
\resizebox{\linewidth}{!}{
\begin{tabular}{lc|cc|cc|cc|cc} 
\toprule
          
\multicolumn{2}{c}{Prediction ($T$)$\rightarrow$}& \multicolumn{2}{c|}{96}& \multicolumn{2}{c|}{192} & \multicolumn{2}{c|}{336} & \multicolumn{2}{c}{720}\\ 
\midrule
$\mathcal{D}$& $L$ & MSE & MAE & MSE & MAE & MSE & MAE& MSE&MAE\\
\midrule
\multirow{3}{*}{\rotatebox{90}{Traffic}}& 192&0.362&0.252&0.386&0.262&0.402&0.270&0.431&0.288\\
& 336& 0.355&0.249&0.378&0.259&0.391&0.266&0.418&0.283\\
& 720&0.348&0.249&0.364&0.255&0.376&0.263&0.410&0.281\\
\midrule
\multirow{3}{*}{\rotatebox{90}{Elec.}}& 192&0.135&0.230&0.167&0.258&0.176&0.269&0.213&0.302\\
& 336&0.133&0.225&0.160&0.255&0.172&0.268&0.211&0.303\\
& 720&0.133&0.225&0.160&0.257&0.167&0.269&0.204&0.300\\
\midrule
\multirow{3}{*}{\rotatebox{90}{ETTm2}}& 192&0.170&0.252&0.230&0.294&0.273&0.325&0.351&0.376\\
& 336&0.165&0.254&0.223&0.291&0.264&0.323&0.345&0.375\\
& 720&0.163&0.253&0.222&0.295&0.265&0.325&0.336&0.376\\
\bottomrule
\end{tabular}
}
\end{table}

\begin{figure}[hbt]
    \centering
    \includegraphics[width=.5\linewidth]{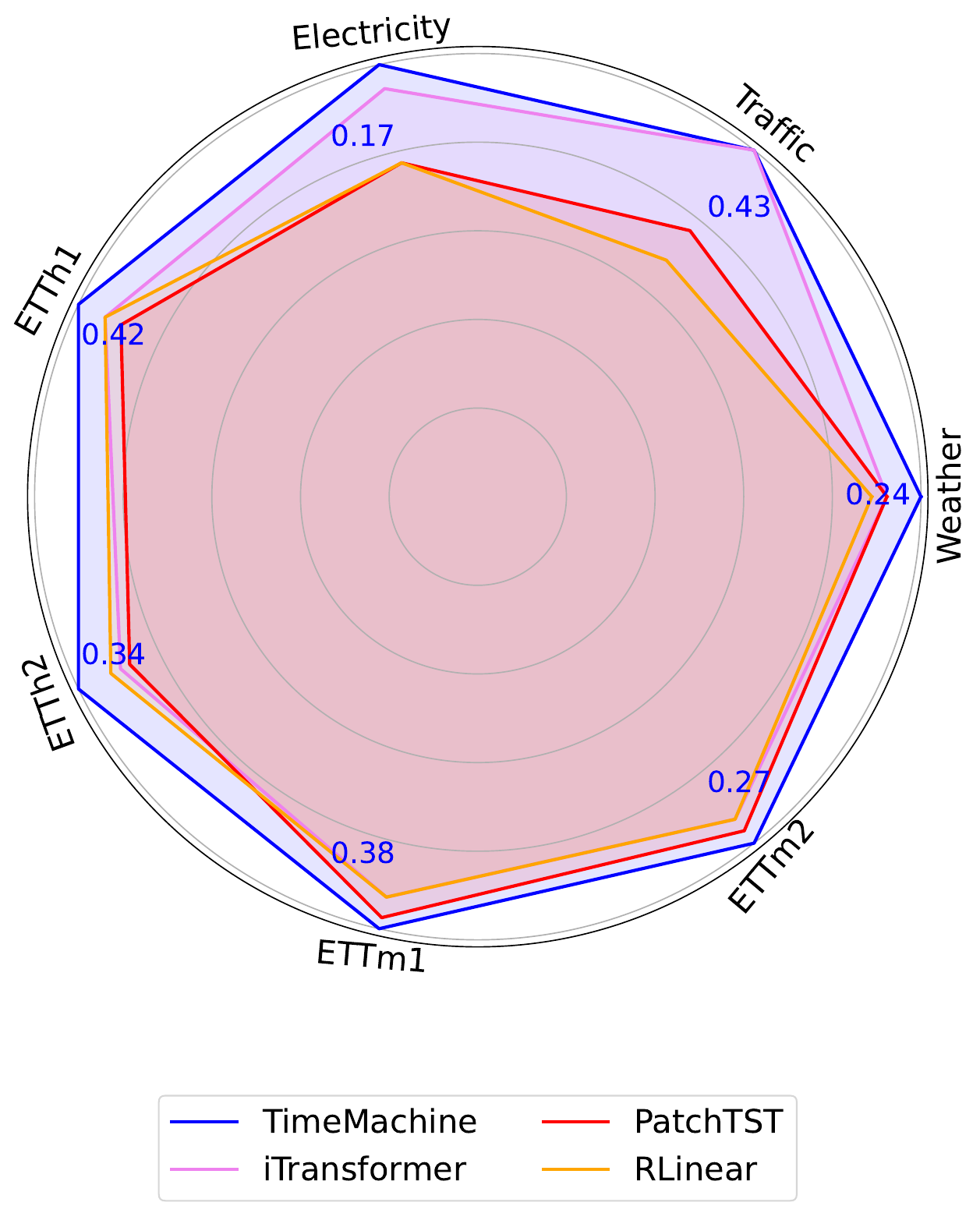}
    \caption{Average MSE comparison of TimeMachine and SOTA baselines with $L=96$. The circle center represents the maximum possible error. Closer to the boundary indicates better performance.}
    \vspace{4mm}
    \label{fig:radar}
\end{figure}

\begin{figure}[htb]
    \centering
   \subfloat[Electricity]{{\includegraphics[width=0.45\linewidth]{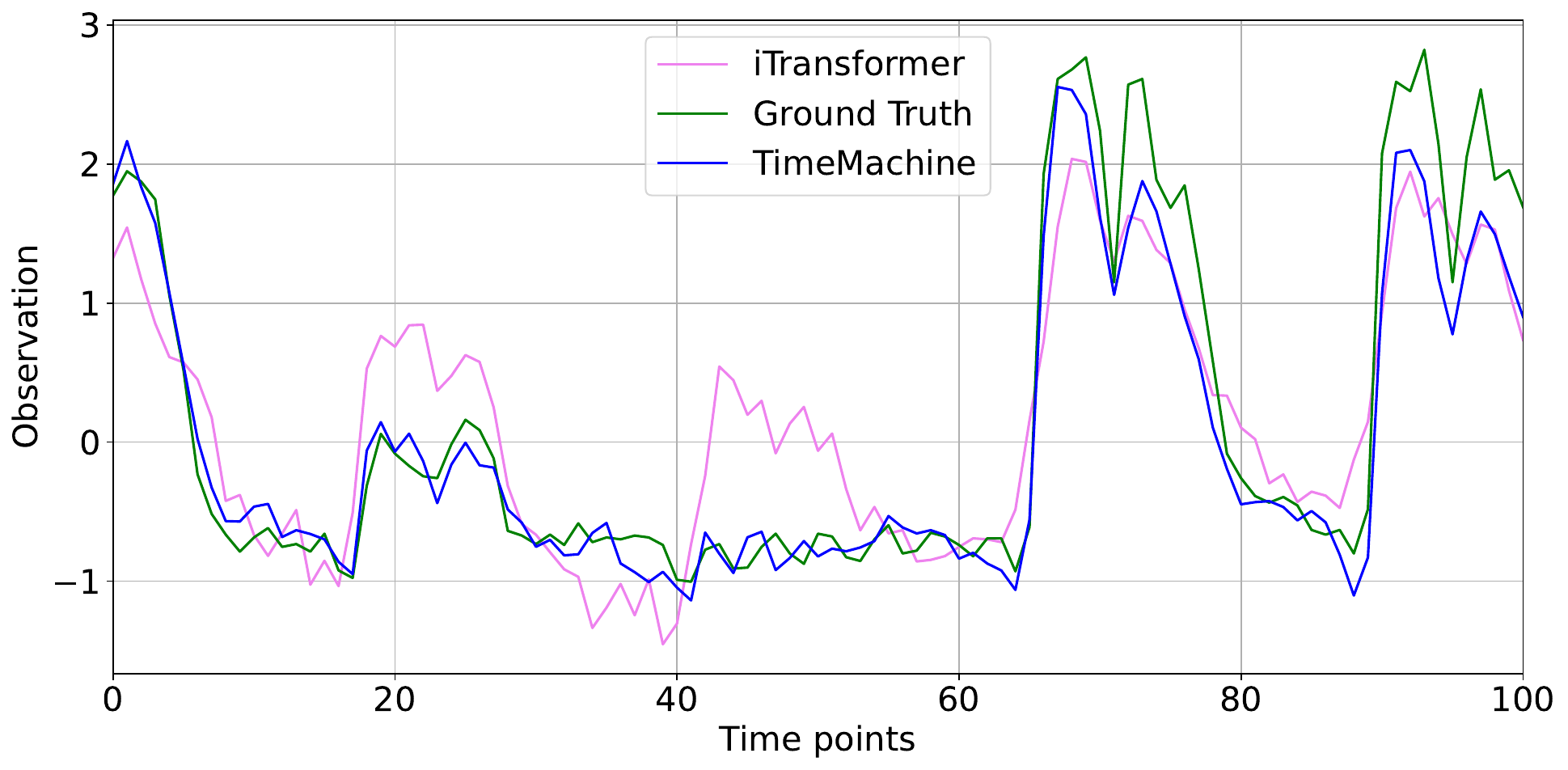} }}%
\qquad
      \subfloat[Traffic]{{\includegraphics[width=0.45\linewidth]{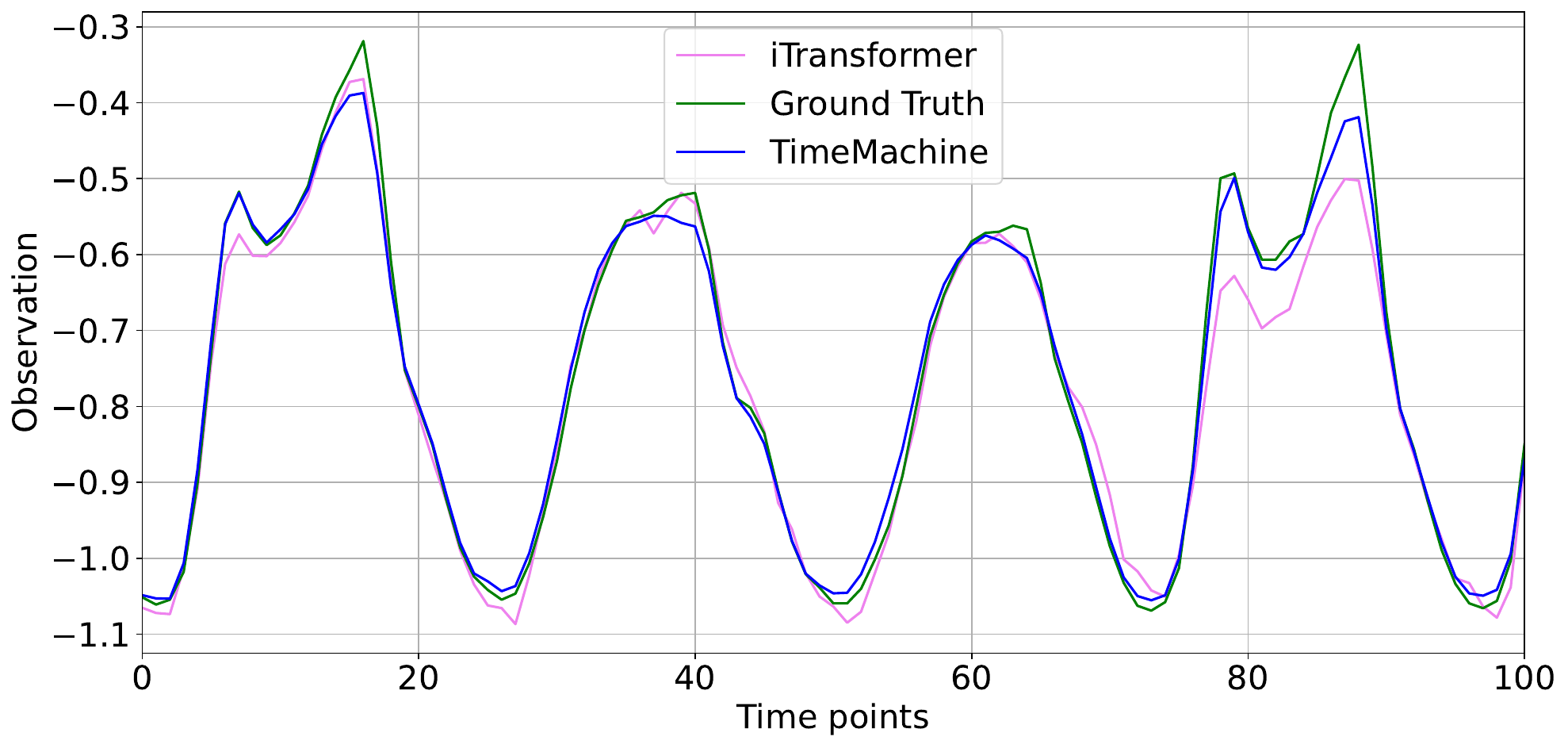} }}%
    \caption{Qualitative comparison between TimeMachine and the second-best method (Table~\ref{tab:result}) on a test set example with L=96, T=720, and a randomly selected channel (best viewed zoomed-in).}
    \vspace{5mm}
    
    \label{fig:qualitative_result}%
\end{figure}

\begin{figure}[hbt]
    \centering
    \subfloat[Traffic]{{\includegraphics[width=0.7\linewidth]{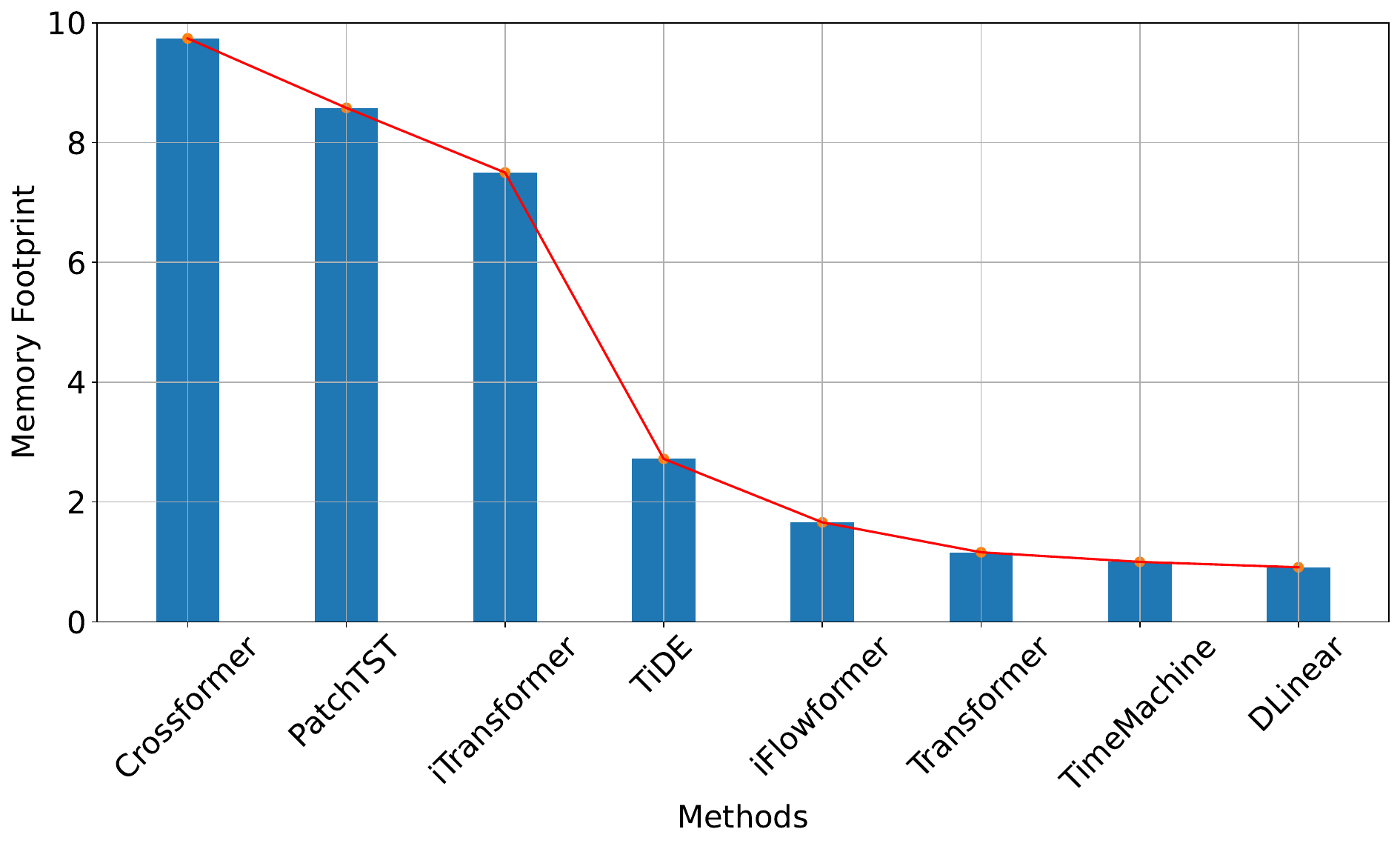} }}%
    \qquad
   \subfloat[Weather]{{\includegraphics[width=0.7\linewidth]{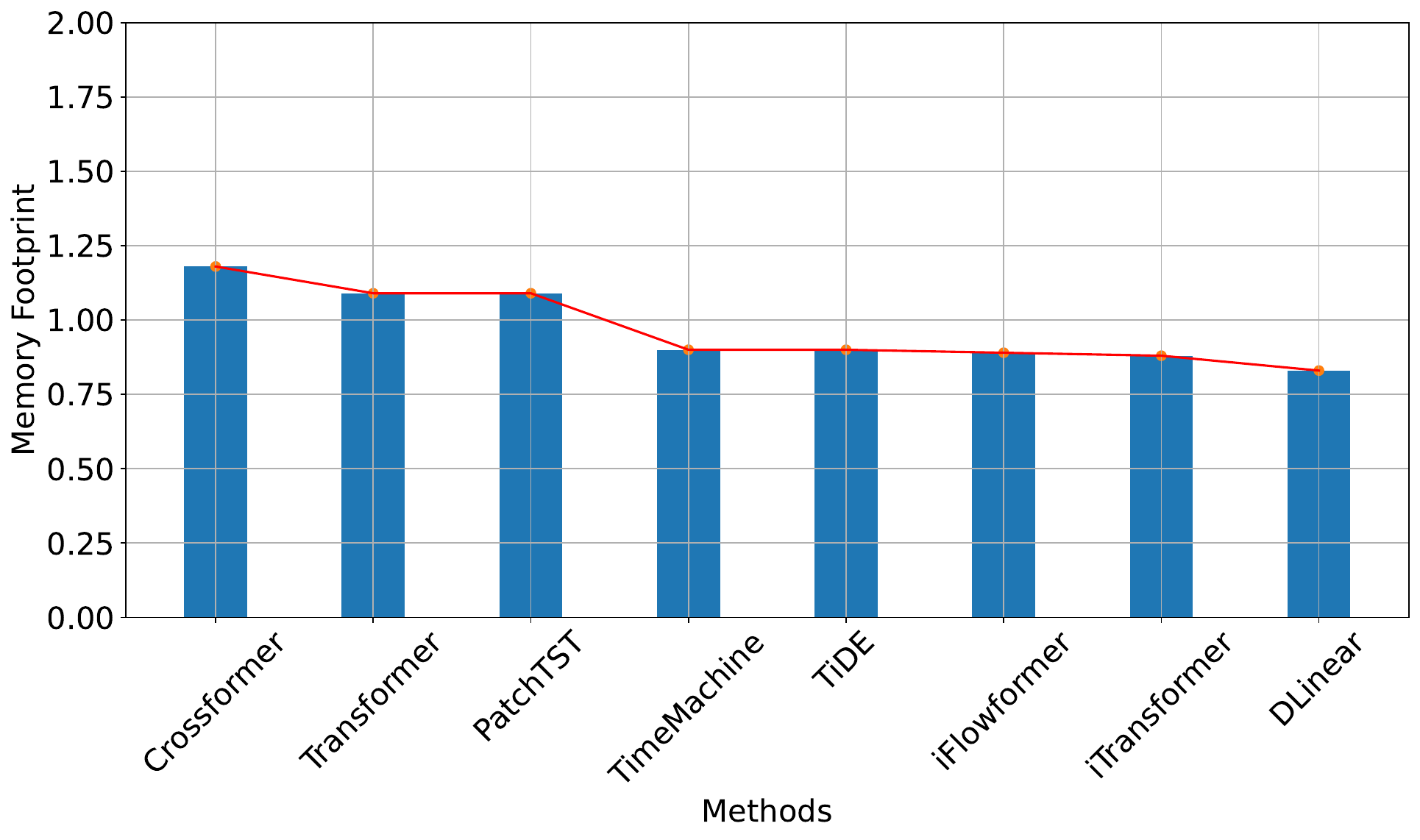} }}%
    \caption{Memory footprint (in GB) for Traffic (with 862 channels) and Weather (with 21 channels) following iTransformer~\citep{liu2024itransformer}.}
    % \vspace{4mm}
    \label{fig:memory}%
\end{figure}

Following iTransformer~\citep{liu2024itransformer}, Figure~\ref{fig:radar} demonstrates the normalized percentage gain of TimeMachine with respect to three other SOTA methods, indicating a clear improvement over the strong baselines.
In addition to the general performance comparison using MSE and MAE metrics, we also compare the memory footprints and scalability of our method against other baselines in Figure~\ref{fig:memory}. We measured the GPU memory utilization of our method and compared it against other baselines, with their results taken from the iTransformer~\citep{liu2024itransformer} paper. To ensure a fair comparison, we also included  Flowformer~\citep{wu2022flowformer} and vanilla Transformer~\citep{vaswani2017attention}, and set the experimental settings for our method similar to those of iTransformer. 

The results clearly show very small memory footprints compared to SOTA baselines. Specifically for Traffic, our method consumes a very similar amount of memory to the DLinear~\citep{dlinear} method. Moreover, our method is capable of handling longer look-back windows with a relatively linear increase in the number of learnable parameters, as demonstrated in Supplementary Figure~\ref{fig:scale} for two datasets. This is due to the robustness of our method, where $E_1$ is only dependent on the input sequence length $L$, and the rest of the networks are relatively independent of $L$,  leading to a highly scalable model.

\subsection{Qualitative Results}
Figure~\ref{fig:qualitative_result} and supplementary Figure~\ref{fig:qualitative_result_s} demonstrate TimeMachine's effectiveness in visual comparison. It is evident that TimeMachine can follow the actual trend in the predicted future time horizon for the test set. In the case of the Electricity dataset, there is a clear difference between the performance of TimeMachine and iTransformer. For the Traffic dataset, although both iTransformer and Timemachine's performance align with the ground truth, in the range approximately between 75-90, TimeMachine's performance is more closely aligned with the ground truth compared to iTransformer. For better visualization, we demonstrated a window of 100 predicted time points. 

\section{Hyperparameter Sensitivity Analysis and Ablation Study}
In this section, we conducted experiments on various hyper-parameters, including training and method-specific parameters. For each parameter, we provided experimental justification based on the achieved results. While conducting an ablation experiment on a parameter, other parameters were kept fixed at their default values, ensuring a clear justification for that specific parameter.
\label{sec:ablation}
\subsection{Effect of MLPs' Parameters ($n_1,n2$)}
\begin{figure}[th]
    \centering
    \subfloat[ETTh1]{{\includegraphics[width=0.7\linewidth]{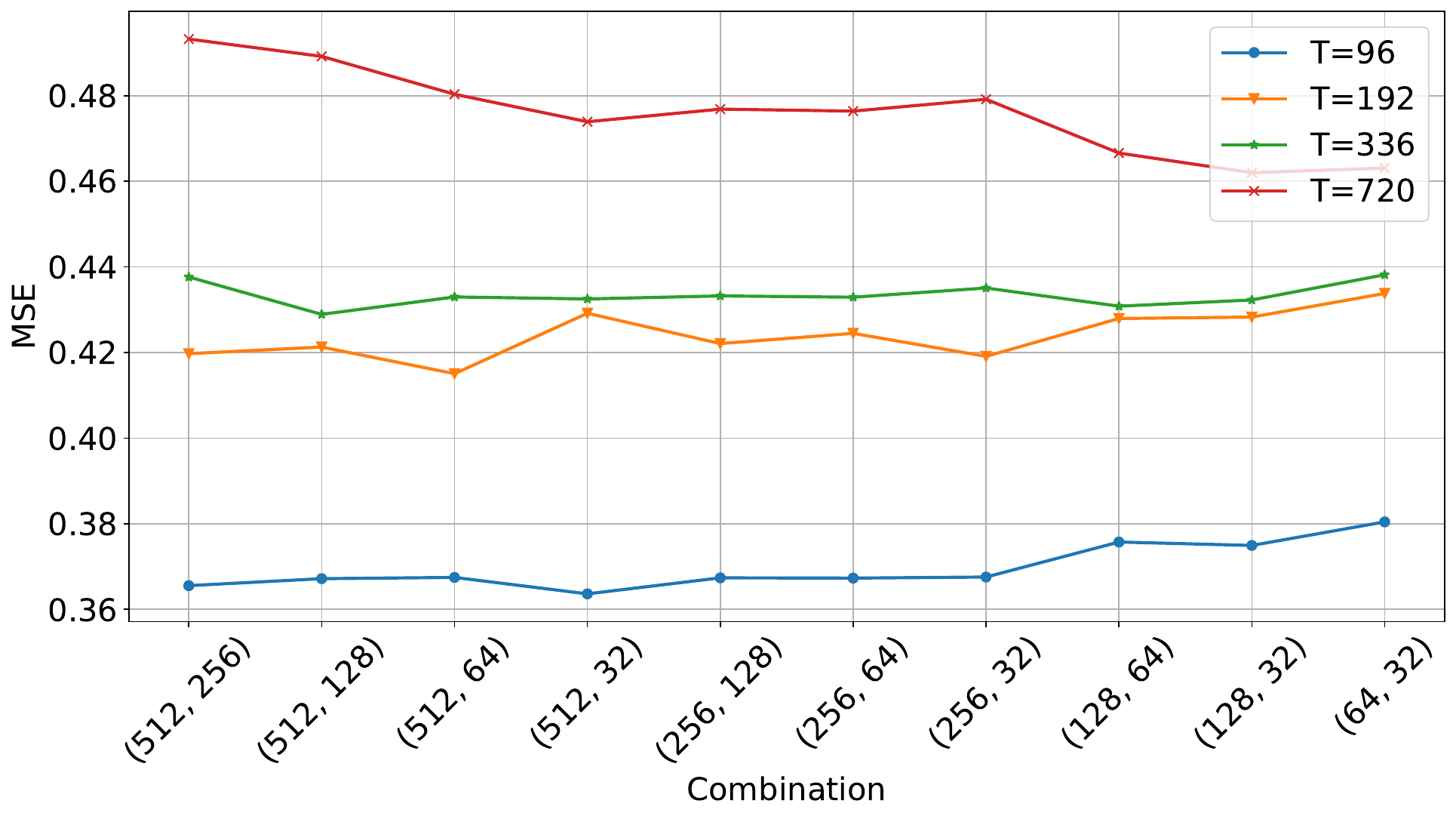} }}%
    \qquad
   \subfloat[ETTh2]{{\includegraphics[width=0.7\linewidth]{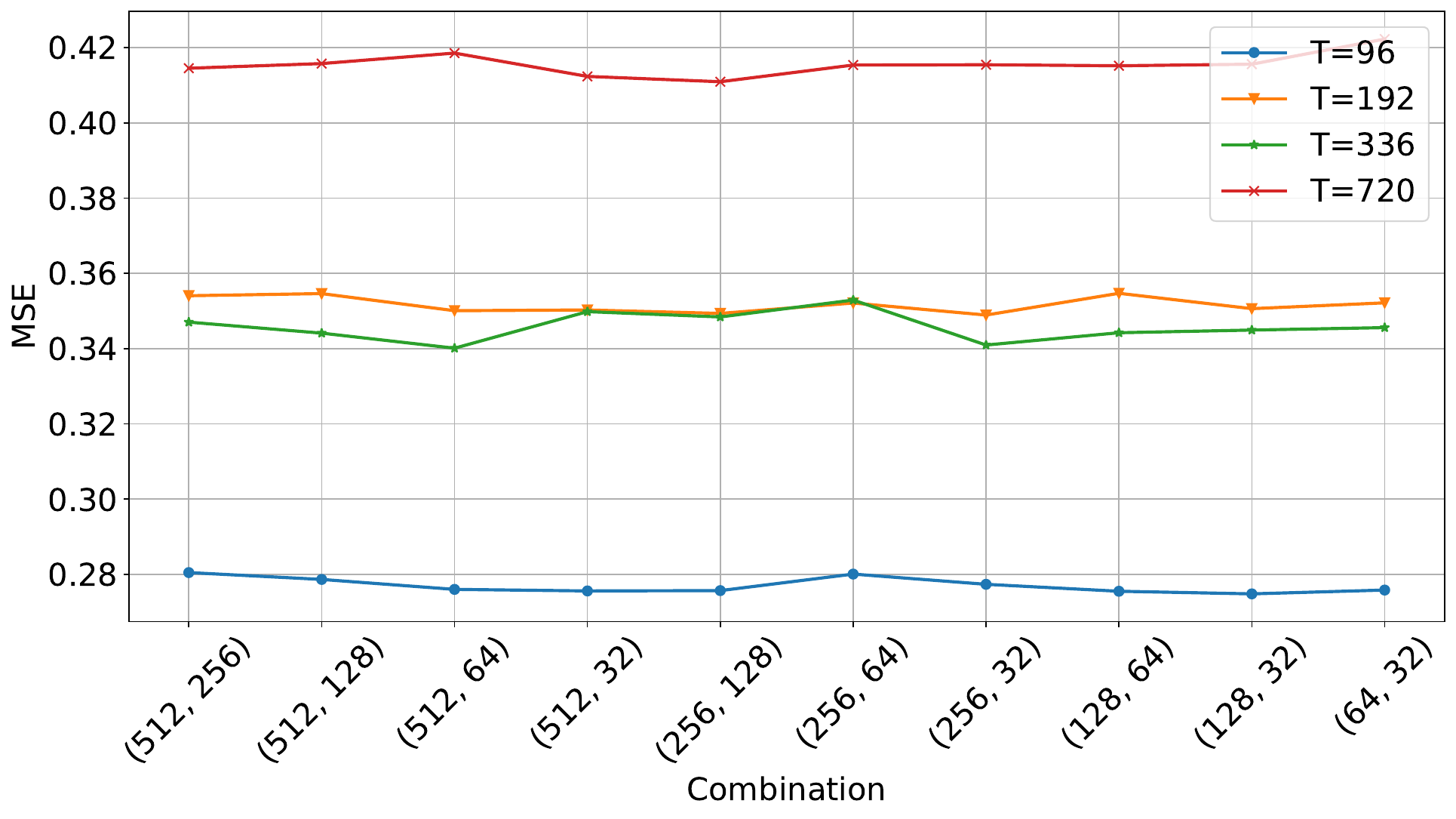} }}%
   
    \caption{MSE comparison with combinations of $n_1$ and $n_2$ for input sequence length $L=96$ for the ETTh1 and ETTh2 datasets.}
    \vspace{5mm}
    \label{fig:emb_ablation}%
\end{figure}

As demonstrated in Figure~\ref{fig:method}, we have two stages of compression with two MLPs $E_1, E_2$ of output dimensions $n_1$ and $n_2$, respectively, and $P_1$ performing an expansion by converting $n_2 \rightarrow n_1$. Since several strong baseline methods, e.g., DLinear, leverage mainly MLPs, we aim at understanding the effect of MLPs on performance. To this end, we explored 10 different combinations from $\{512, 256, 128, 64, 32\}$ and demonstrated the performance with MSE for two datasets (ETTh1, ETTh2) in Figure~\ref{fig:emb_ablation}. These figures show that our method is not heavily dependent on the MLPs. Rather, we can see more improvement with very small MLPs for $T=720$ with the ETTh1 dataset and mostly stable performance on the ETTh2 dataset. 
\subsection{Sensitivity of Dropouts}
In our model (Figure~\ref{fig:method}), we include two dropouts after processing the signals from $E_1$ and $E_2$. These dropouts are necessary, especially for datasets with a small number of channels, e.g., ETTs. Supplementary Figure~\ref{fig:dropout_ablation} shows the effect of dropouts on both ETTh1 and ETTh2 datasets. As expected, too low or too high dropout rates are not helpful. To maintain balance, we set the dropout rates to $0.7$ for both datasets while tuning other variations for the rest.

\subsection{Ablation of Residual Connections}
Studies have shown the effectiveness of residual connection, including models using SSMs~\citep{mambatab} and CNNs~\citep{resnet}. In this section, we justify the two residual connections in our architecture: one from $E_2$ to the output of the two inner Mambas, and the other from $E_1$ to the output of $P_1$. Both of them use element-wise additions and help stabilize training and reduce overfitting, especially for the smaller datasets with channel independence. Supplementary Table~\ref{tab:res_connections} provides experimental justification, where the Res. column indicates the presence  (\cmark) or absence (\xmark) of residual connections. We observe clear improvement on both datasets when residual connections are used. This motivated us to include residual connections in our architecture, and all results presented in Tables~\ref{tab:result} and~\ref{tab:result_long_l} incorporate these connections.

\subsection{Effects of Mambas' Local Convolutional Width}
In addition to experimenting with the different components of our architecture (Figure~\ref{fig:method}), we also investigated the effectiveness of Mamba parameters. For example, we tested two variations of local convolutional kernel widths ($2$ and $4$) for the Mambas and found that a kernel width of $2$ yields more promising results compared to 4. Therefore, we set the default kernel width to $2$ for all datasets and Mambas.
\begin{table}[t]
\centering
\vspace{1mm}
\caption{Ablation results on the local convolution width with $L=96$.}
\label{tab:d_conv}
\setlength{\tabcolsep}{2pt}
\medskip

\begin{tabular}{lc|cc|cc|cc|cc} 
\toprule
          
\multicolumn{2}{c}{Prediction ($T$)$\rightarrow$}& \multicolumn{2}{c|}{96}& \multicolumn{2}{c|}{192} & \multicolumn{2}{c|}{336} & \multicolumn{2}{c}{720}\\ 
\midrule
$\mathcal{D}$&d\_conv& MSE & MAE & MSE & MAE & MSE & MAE& MSE&MAE\\
\midrule
\multirow{2}{*}{ETTh1}&4&0.365&0.389&0.419&0.418&0.439&0.424&0.465&0.457\\
&2&\bf0.364&\bf0.387&\bf0.415&\bf0.416&\bf0.429&\bf0.421&\bf0.458&\bf0.453\\
\midrule
\multirow{2}{*}{ETTh2}&4&0.275&\bf0.333&\bf0.347&0.383&0.350&0.382&0.411&0.433\\
&2&\bf0.275&0.334&0.349&\bf0.381&\bf0.340&\bf0.381&\bf0.411&\bf0.433\\
\bottomrule
\end{tabular}

\end{table}

\subsection{Ablation on State Expansion Factor of Mambas}
\begin{figure}[t]
    \centering
    \subfloat[ETTh1]{{\includegraphics[width=0.7\linewidth]{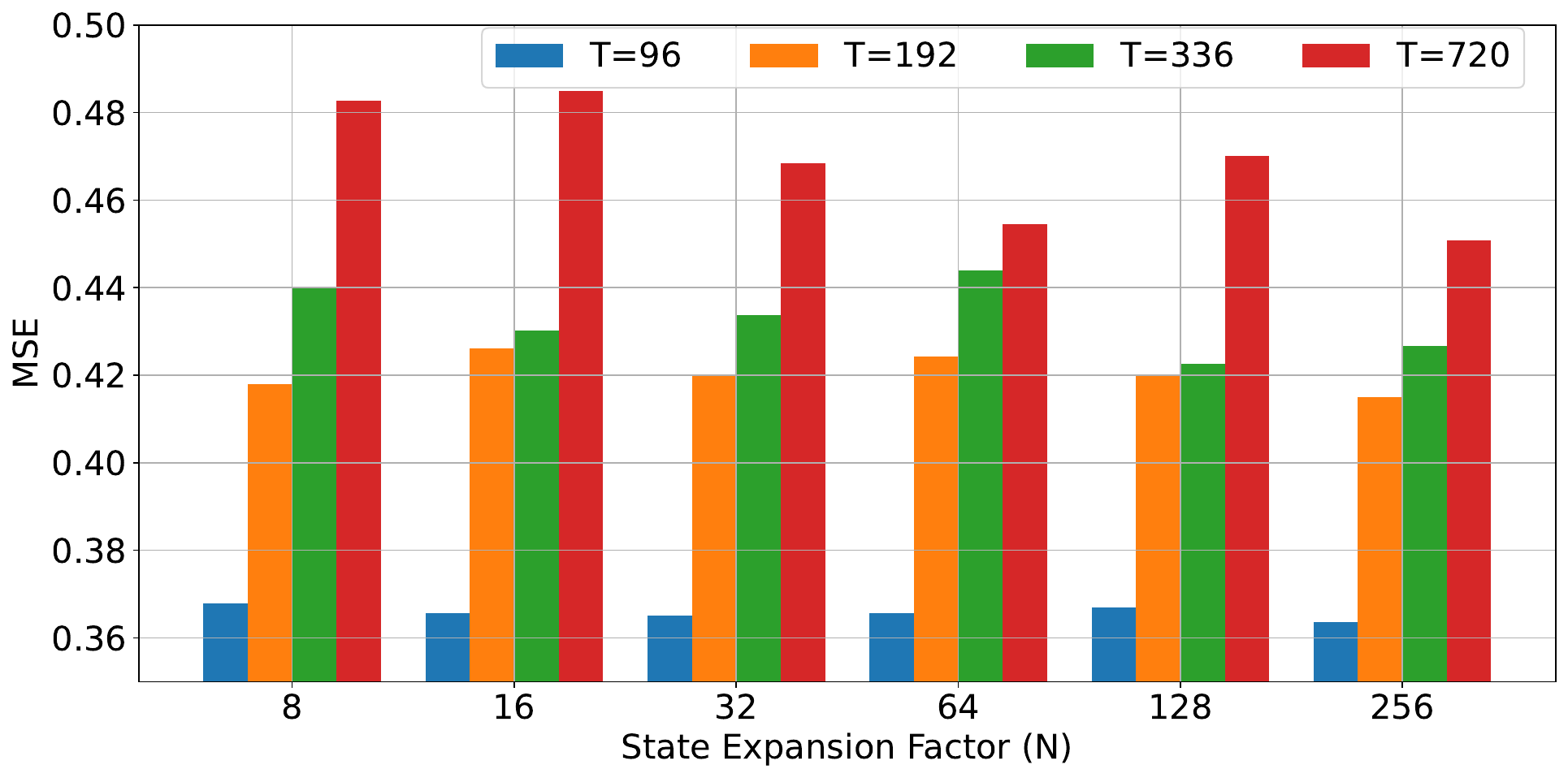} }}%
    \qquad
   \subfloat[ETTh2]{{\includegraphics[width=0.7\linewidth]{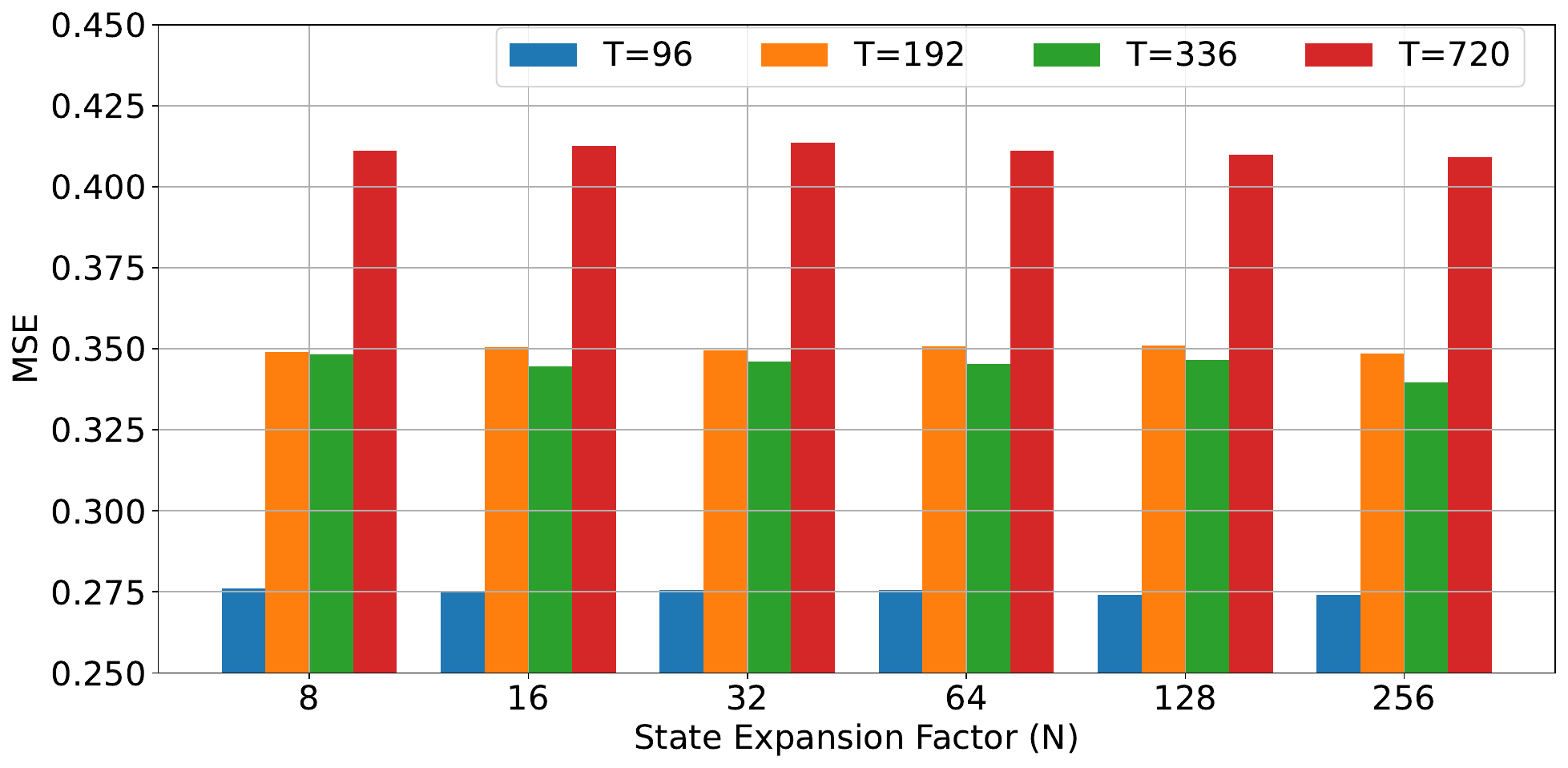} }}%
   
    \caption{MSE versus the state expansion factor ($N$) with the input sequence length $L=96$.}
    \vspace{5mm}
    \label{fig:dstate_ablation}%
\end{figure}

The SSM state expansion factor ($N$) is another crucial parameter of Mamba. We ablate this parameter from a very small value of 8 up to the highest possible value of 256. Figure~\ref{fig:dstate_ablation} demonstrates the effectiveness of this expansion factor while keeping all other parameters fixed. With a higher state expansion factor, there is a certain chance of performance improvement for varying prediction lengths. Therefore, we set $N=256$  as the default value for all datasets, and the results in Tables~\ref{tab:result} and ~\ref{tab:result_long_l} contain the TimeMachine's performance with this default value.
\subsection{Ablation on Mamba Dimension Expansion Factor}
We also experimented with the dimension expansion factor ($E$) of the Mambas, which is used to expand the input dimension, with results shown in Supplementary Figure~\ref{fig:expand_ablation}. Increasing the block expansion factor does not lead to consistent improvements in performance. Instead, higher expansion factors come with a heavy cost in memory and training time. Therefore, we set this value to 1 by default in all Mambas and report the results in Tables~\ref{tab:result} and ~\ref{tab:result_long_l}.

In addition to these sensitivity analyses, we also demonstrated performance comparison between 1 and 2 levels in Supplementary Table~\ref{tab:level}. Considering a balance between performance and memory footprint, we used two levels.

\section{Strengths and Limitations}
TimeMachine outperforms numerous baselines, including transformer-based methods, across benchmark datasets and additionally demonstrates memory efficiency and stable performance across varying look-back and prediction lengths. Unlike transformer-based methods that have quadratic complexity, our method has linear complexity. While TimeMachine achieves top-ranked performance in most cases, it ranks second on the Weather dataset with small $T$, highlighting an area for future improvement. Moreover, as shown in Figure~\ref{fig:qualitative_result}, there is potential for enhancing alignment with ground truth.
\section{Conclusion}
This paper introduces TimeMachine, a novel model that captures long-term dependencies in multivariate time series data while maintaining linear scalability and small memory footprints. By 
%exploiting the multi-scale properties of time series data and 
leveraging an integrated quadruple-Mamba architecture to predict with rich global and local contextual cues at multiple scales, TimeMachine unifies channel-mixing and channel-independence situations, enabling accurate long-term forecasting. Extensive experiments demonstrate the model's superior performance in accuracy, scalability, and memory efficiency compared to state-of-the-art methods. Future work will explore TimeMachine's application in a self-supervised learning setting. 
%%%%%%%%%%%%%%%%%%%%%%%%%%%%%%%%%%%%%%%%%%%%%%%%%%%%%%%%%%%%%%%%%%%%%%%%

%%% Use this environment to include acknowledgements (optional).
%%% This will be omitted in doubleblind mode.
\clearpage
\begin{ack}
This research is supported in part by the NSF under Grants 2327113 and 2433190 and the NIH under Grants R21AG070909, P30AG072946, and R01HD101508-01.
We would like to thank the University of Kentucky Center for Computational Sciences and Information Technology Services Research Computing for their support and use of the Lipscomb Compute Cluster and associated research computing resources.
\end{ack}

%%%%%%%%%%%%%%%%%%%%%%%%%%%%%%%%%%%%%%%%%%%%%%%%%%%%%%%%%%%%%%%%%%%%%%%%

%%% Use this command to include your bibliography file.

\bibliography{2.mybibfile}
\clearpage
\setcounter{table}{0}
\setcounter{figure}{0}
\setcounter{page}{1}
\nolinenumbers

\captionsetup[table]{name={Supplementary Table}}
\captionsetup[figure]{name={Supplementary Figure}}
\section*{Supplementary Materials}
\begin{table}[htb]
\centering
\caption{Ablation experiment on the residual connections with input sequence length $L=96$ and $T=\{96,192,336,720\}$.}
\label{tab:res_connections}
\setlength{\tabcolsep}{2pt}
\medskip
\resizebox{\linewidth}{!}{
\begin{tabular}{lc|cc|cc|cc|cc} 
\toprule
          
\multicolumn{2}{c}{Prediction ($T$)$\rightarrow$}& \multicolumn{2}{c|}{96}& \multicolumn{2}{c|}{192} & \multicolumn{2}{c|}{336} & \multicolumn{2}{c}{720}\\ 
\midrule
$\mathcal{D}$& Res.& MSE & MAE & MSE & MAE & MSE & MAE& MSE&MAE\\
\midrule
\multirow{2}{*}{ETTh1}&\xmark&0.366&0.395&0.423&0.425&0.430&0.427&0.474&0.462\\
&\cmark&\bf0.364&\bf0.387&\bf0.415&\bf0.416&\bf0.429&\bf0.421&\bf0.458&\bf0.453\\
\midrule
\multirow{2}{*}{ETTh2}&\xmark&0.281&0.337&0.347&0.386&0.352&0.383&0.415&0.435\\
&\cmark&\bf0.275&\bf0.334&\bf0.349&\bf0.381&\bf0.340&\bf0.381&\bf0.411&\bf0.433\\
\bottomrule
\end{tabular}
}
\end{table}

\begin{figure}[htb]
    \centering
    \subfloat[ETTh1]{{\includegraphics[width=0.7\linewidth]{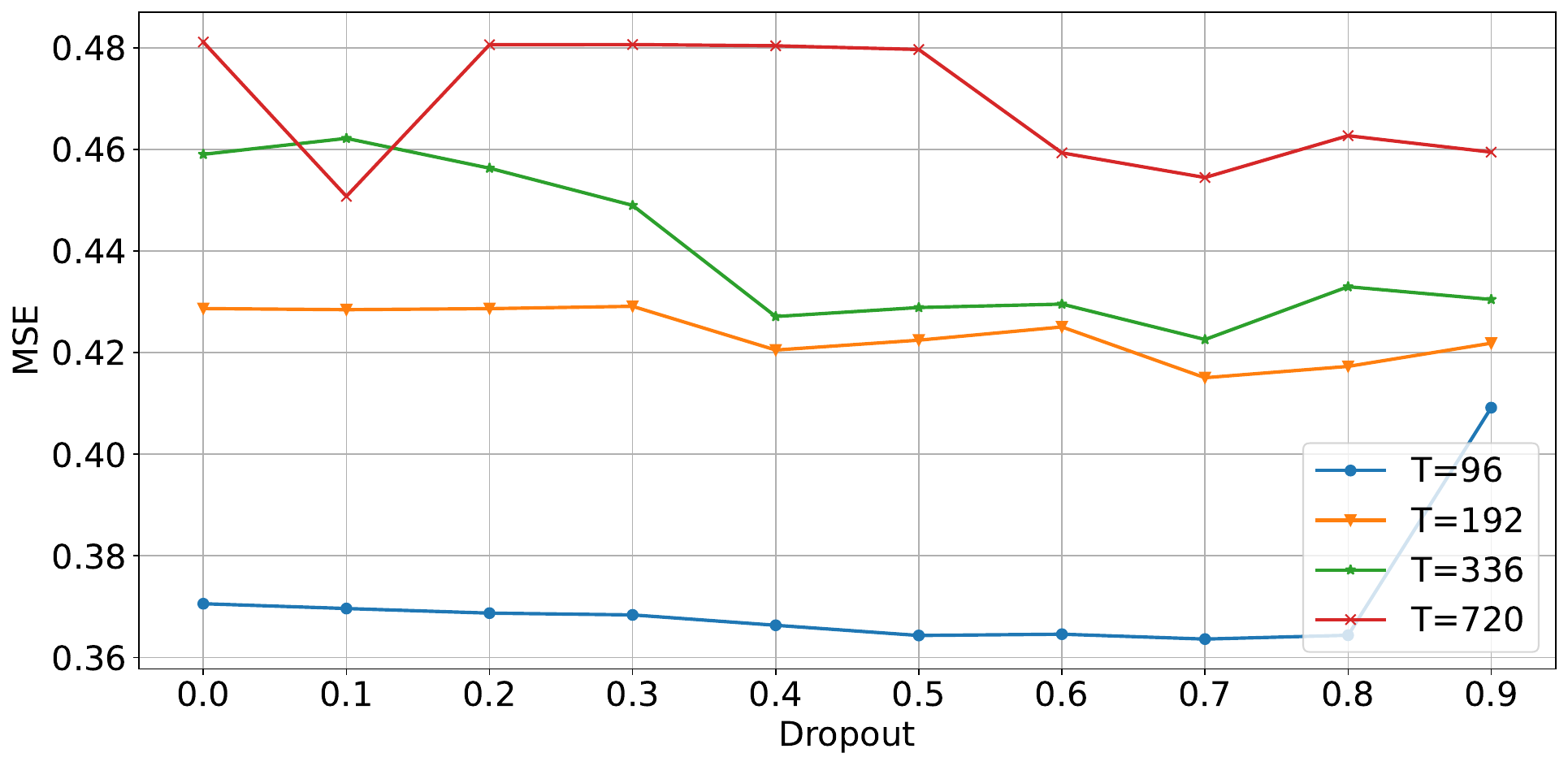} }}%
    \qquad
   \subfloat[ETTh2]{{\includegraphics[width=0.7\linewidth]{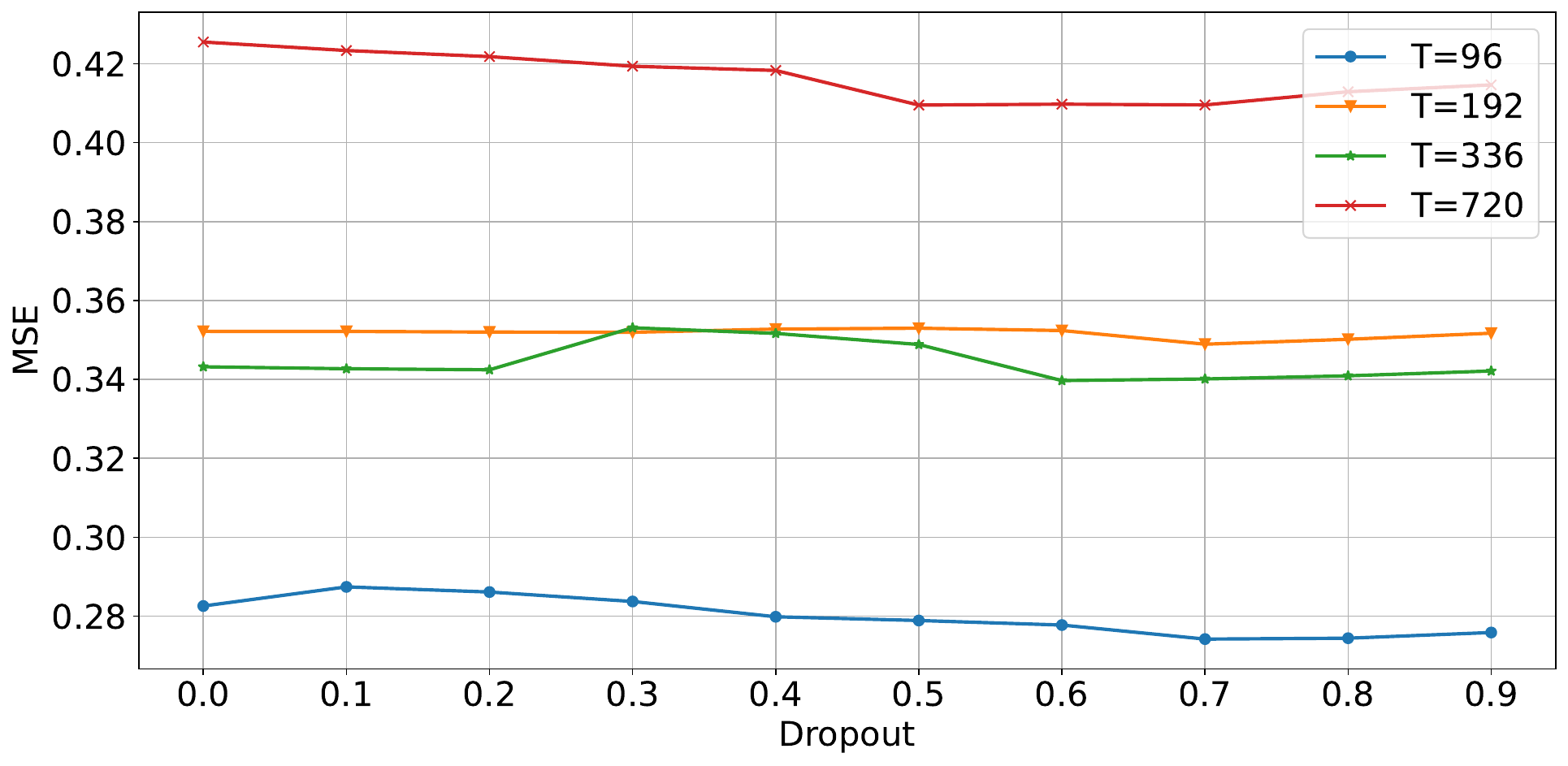} }}%
   
    \caption{Performance (MSE) comparison concerning a diverse range of dropouts with input sequence length $L=96$.}
    \vspace{4mm}
    \label{fig:dropout_ablation}%
\end{figure}

\begin{figure}[t]
    \centering
    \subfloat[ETTm1]{{\includegraphics[width=0.7\linewidth]{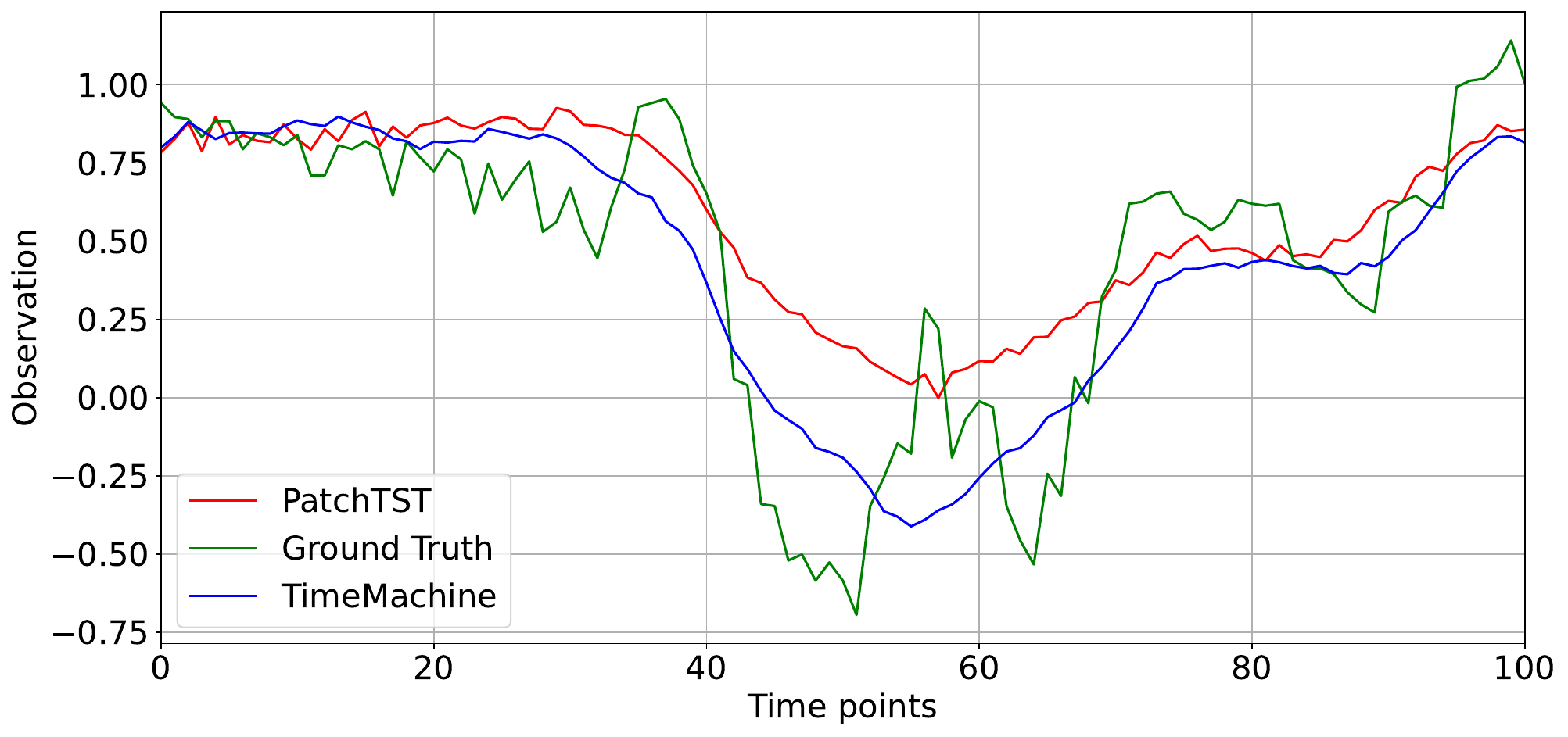} }}%
    \qquad
   \subfloat[ETTm2]{{\includegraphics[width=0.7\linewidth]{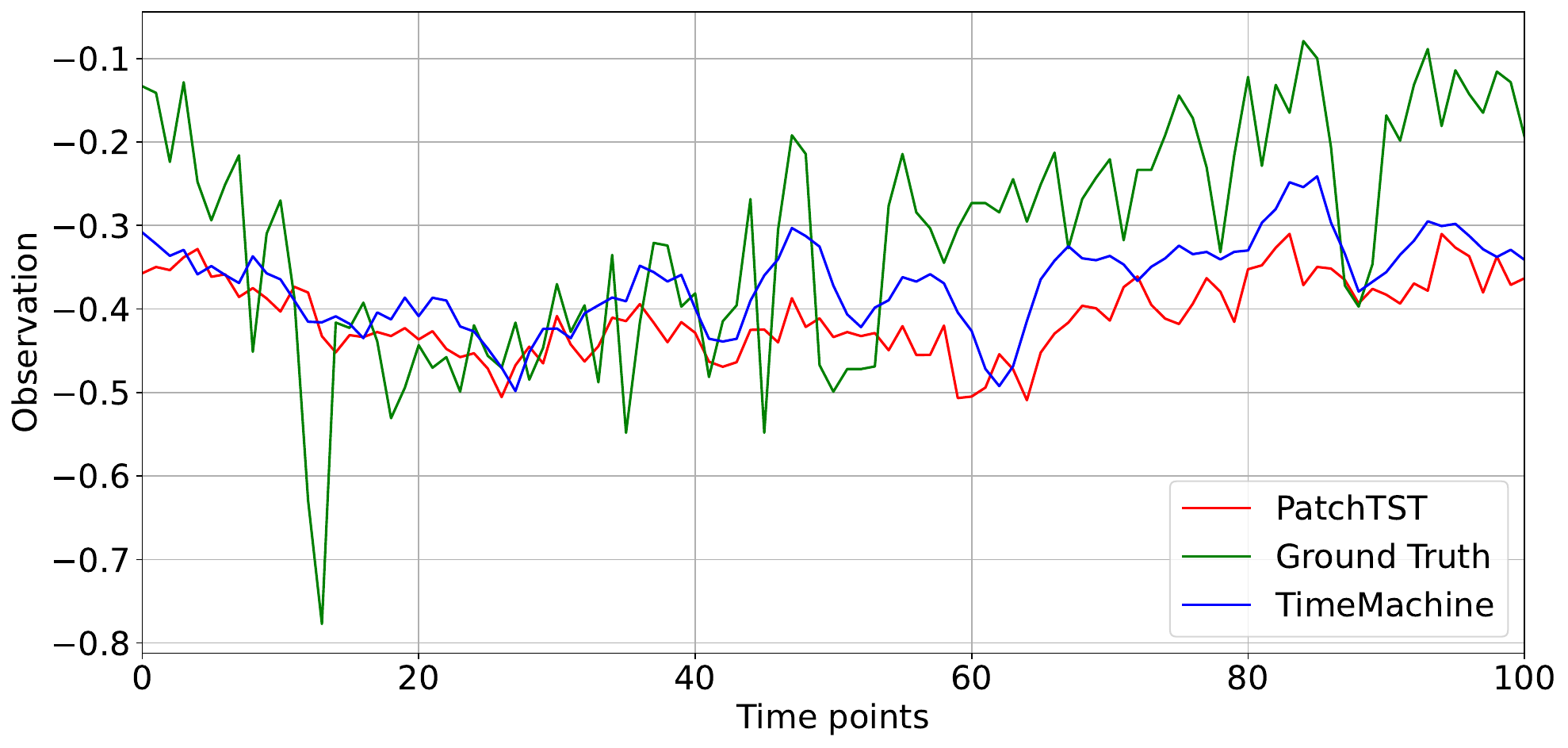} }}%
    \caption{Qualitative comparison between TimeMachine and second-best-performing methods from Table ~\ref{tab:result}. Visualization is provided for the test set with $L=96$ and $T=720$ with a randomly selected channel and a window frame of $100$ time points.}%
    \label{fig:qualitative_result_s}%
\end{figure}
\begin{table}[htb]
\centering
\caption{Results for the long-term forecasting task with varying input sequence length $L=\{192,336,720\}$ and $T=\{96,192,336,720\}$}
\label{tab:result_long_l_s}
\setlength{\tabcolsep}{2pt}
\medskip
\resizebox{\linewidth}{!}{
\begin{tabular}{lc|cc|cc|cc|cc} 
\toprule
          
\multicolumn{2}{c}{Prediction ($T$)$\rightarrow$}& \multicolumn{2}{c|}{96}& \multicolumn{2}{c|}{192} & \multicolumn{2}{c|}{336} & \multicolumn{2}{c}{720}\\ 
\midrule
$\mathcal{D}$& $L$ & MSE & MAE & MSE & MAE & MSE & MAE& MSE&MAE\\
\midrule
\multirow{3}{*}{\rotatebox{90}{Weather}}&192&0.155&0.204&0.198&0.243&0.241&0.281&0.327&0.336\\
&336&0.151&0.201&0.192&0.240&0.236&0.278&0.318&0.334\\
&720&0.151&0.203&0.195&0.246&0.239&0.285&0.321&0.340\\
\midrule
\multirow{3}{*}{\rotatebox{90}{ETTh1}}&192&0.365&0.386&0.415&0.413&0.406&0.417&0.447&0.459\\
& 336&0.360&0.387&0.398&0.410&0.386&0.411& 0.443&0.457\\
& 720&0.363&0.395&0.402&0.418&0.396&0.420&0.468&0.476\\
\midrule
\multirow{3}{*}{\rotatebox{90}{ETTh2}}&192&0.274&0.334&0.340&0.379&0.327&0.378&0.402&0.432\\
&336&0.267&0.334&0.324&0.375&0.316&0.375&0.392&0.429\\
&720&0.260&0.332&0.314&0.372&0.316&0.377&0.394&0.433\\
\midrule
\multirow{3}{*}{\rotatebox{90}{ETTm1}}&192&0.286&0.337&0.331&0.365&0.354&0.384&0.421&0.421\\
&336&0.286&0.337&0.328&0.364&0.355&0.381&0.408&0.413\\
&720&0.289&0.344&0.334&0.369&0.357&0.382&0.416&0.413\\

\bottomrule
\end{tabular}
}
\end{table}
\begin{figure}[t]
    \centering
    \subfloat[ETTh1]{{\includegraphics[width=0.7\linewidth]{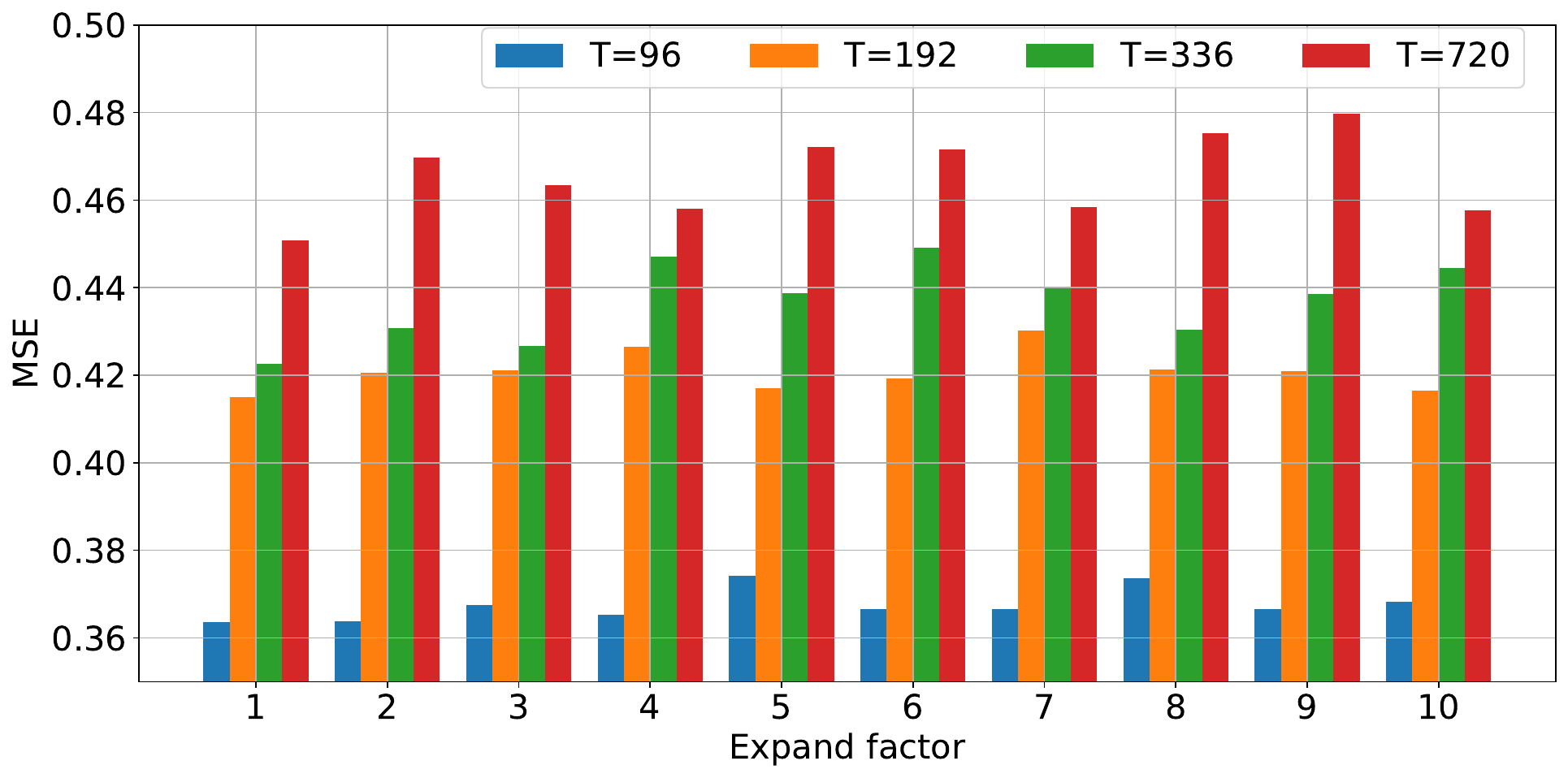} }}%
    \qquad
   \subfloat[ETTh2]{{\includegraphics[width=0.7\linewidth]{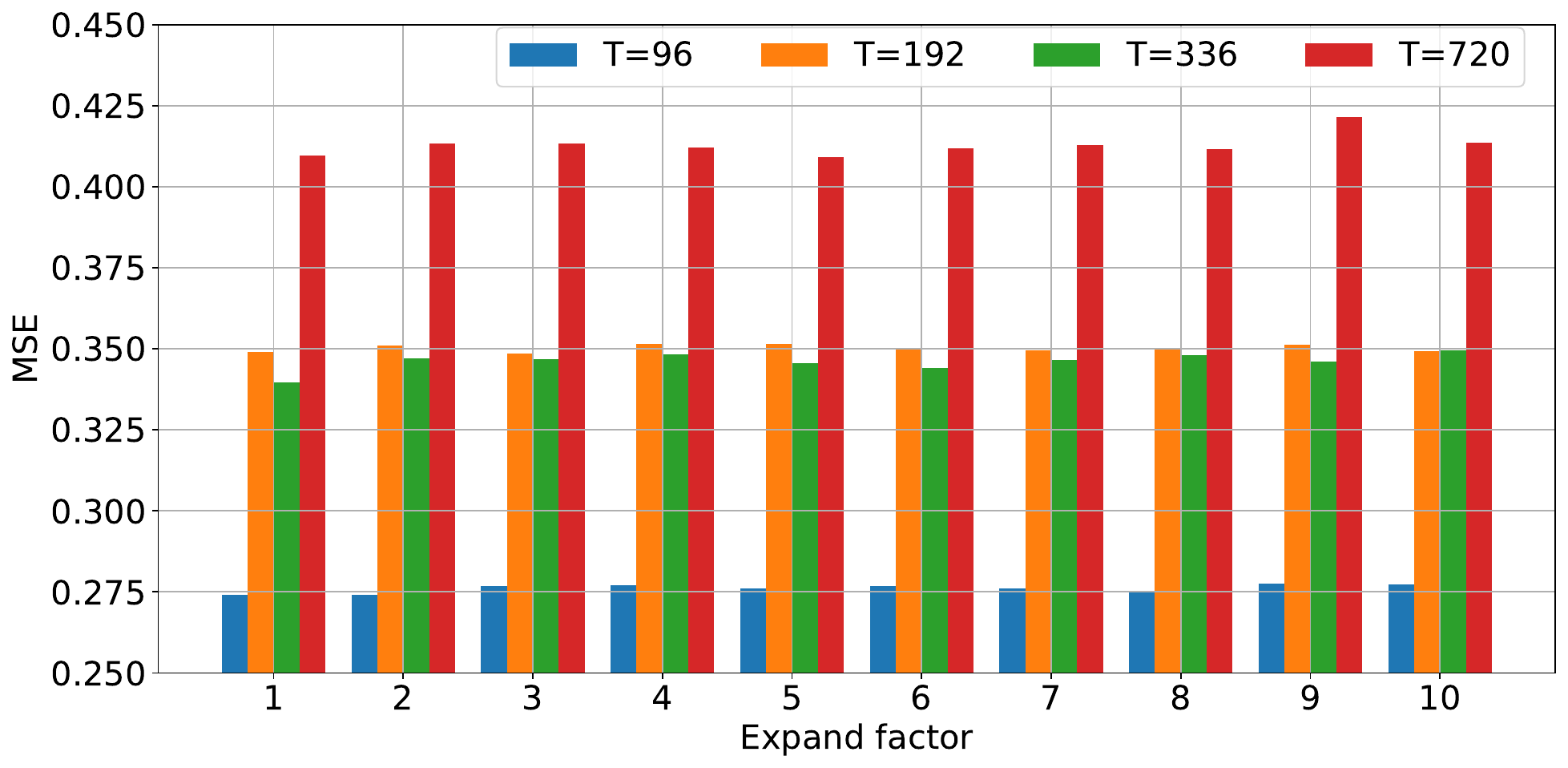} }}%
   
    \caption{Comparative analysis for the expanding factor.}%
    \label{fig:expand_ablation}%
\end{figure}
\begin{figure}[t]
    \centering
    \subfloat[ETTh2]{{\includegraphics[width=0.7\linewidth]{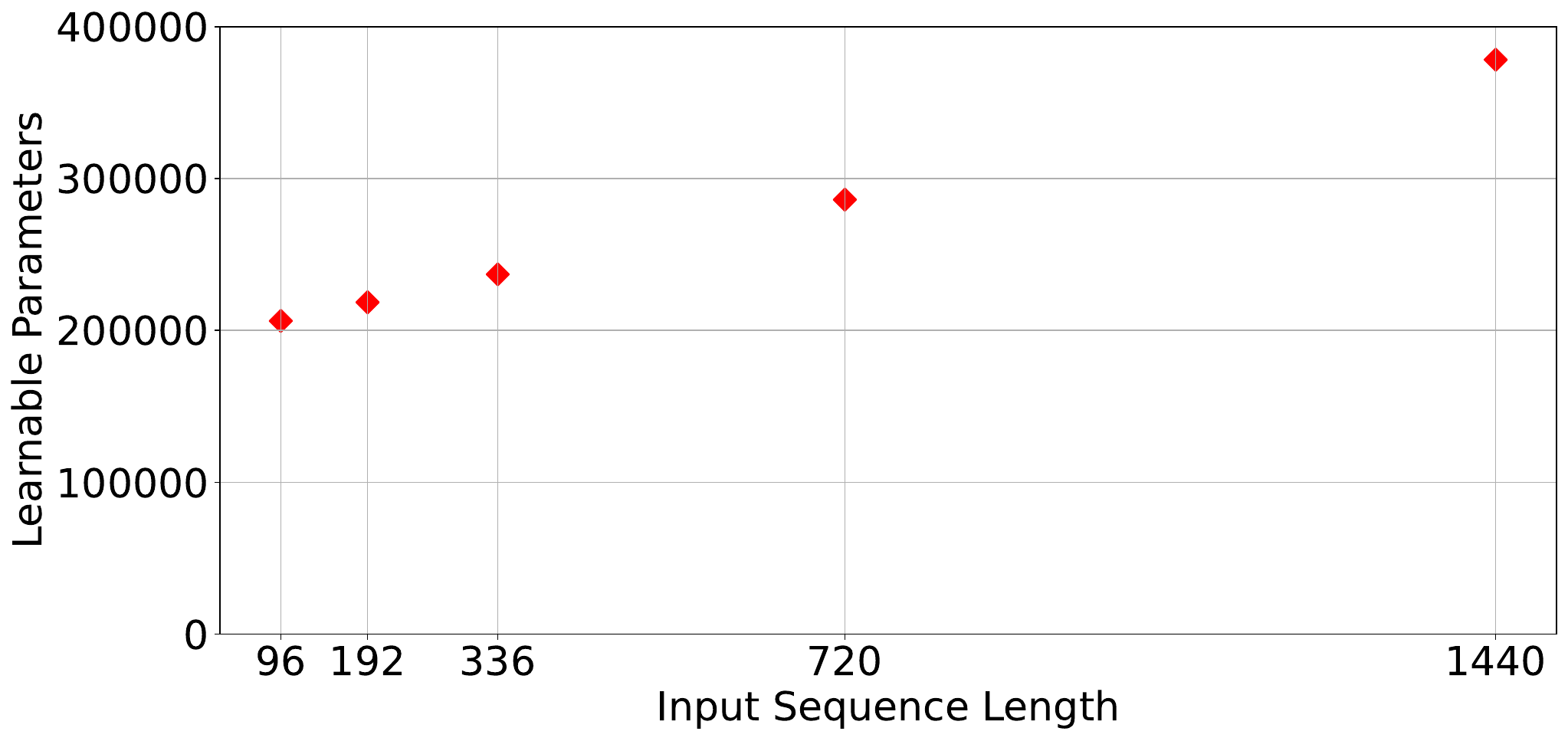} }}%
    \qquad
   \subfloat[Weather]{{\includegraphics[width=0.7\linewidth]{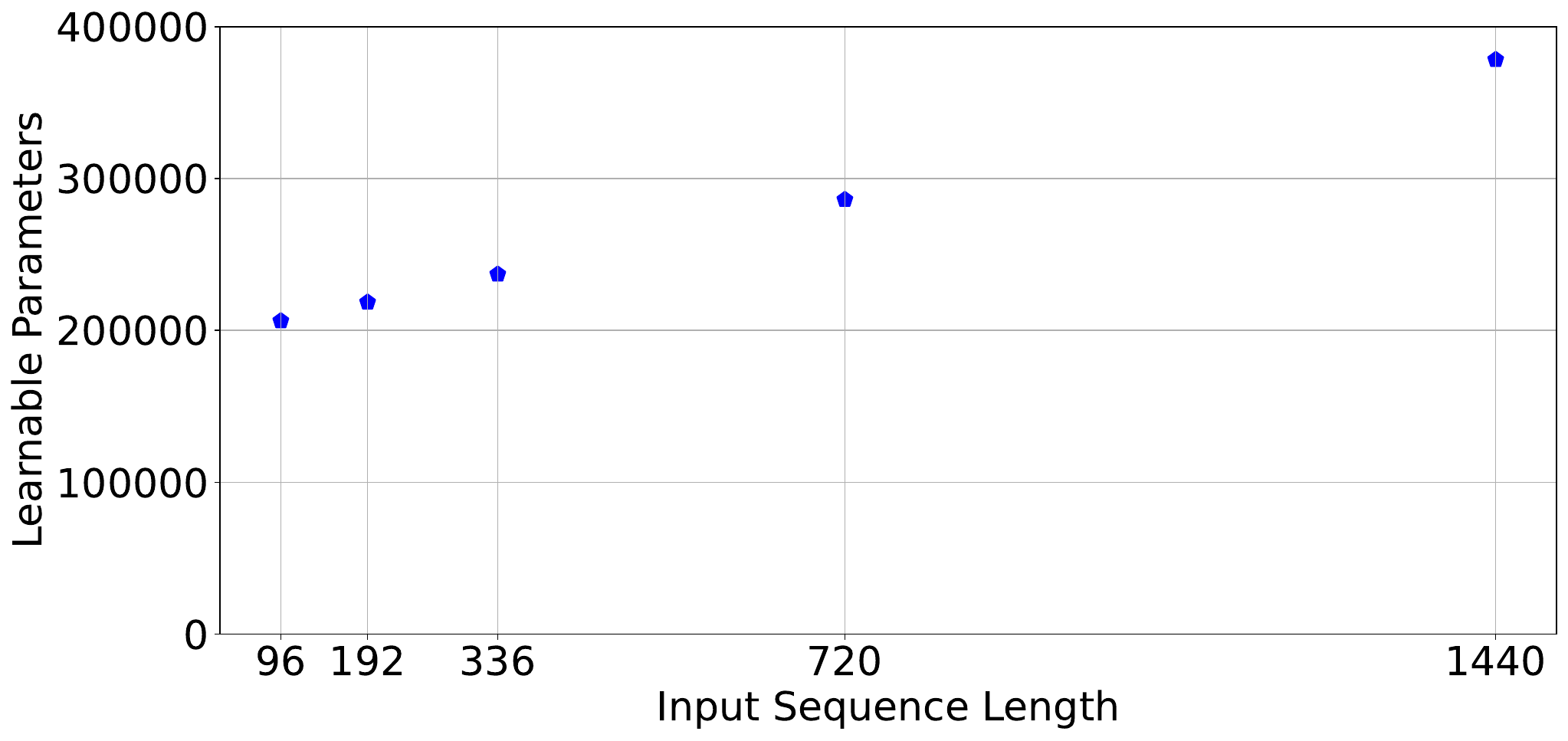} }}%
   
    \caption{Scalability in terms of learnable parameters with respect to look-back window.}%
    \label{fig:scale}%
\end{figure}

\begin{table}[htb]
\centering
\caption{Ablation experiment on the level of TimeMachine with input sequence length $L=96$ and $T=\{96,192,336,720\}$.}
\label{tab:level}
\setlength{\tabcolsep}{2pt}
\medskip
\resizebox{\linewidth}{!}{
\begin{tabular}{lc|cc|cc|cc|cc} 
\toprule
          
\multicolumn{2}{c}{Prediction ($T$)$\rightarrow$}& \multicolumn{2}{c|}{96}& \multicolumn{2}{c|}{192} & \multicolumn{2}{c|}{336} & \multicolumn{2}{c}{720}\\ 
\midrule
$\mathcal{D}$& Level& MSE & MAE & MSE & MAE & MSE & MAE& MSE&MAE\\
\midrule
\multirow{2}{*}{ETTh1}&1&0.367&0.393&0.420&0.418&0.437&0.424&0.460&0.455\\
&2&\bf0.364&\bf0.387&\bf0.415&\bf0.416&\bf0.429&\bf0.421&\bf0.458&\bf0.453\\
\midrule
\multirow{2}{*}{Electricity}&1&0.143&0.237&0.164&0.256&0.175&0.271&0.212&0.303\\
&2&\bf0.142&\bf0.236&\bf0.158&\bf0.250&\bf0.172&\bf0.268&\bf0.207&\bf0.298\\
\bottomrule
\end{tabular}
}
\end{table}

\end{document}